\documentclass[10pt,twocolumn,letterpaper]{article}  

\usepackage{etoolbox}
\newtoggle{cvpr}
\newtoggle{eccv}
\newtoggle{corl}
\toggletrue{cvpr}

\iftoggle{cvpr}{
    \usepackage[pagenumbers]{cvpr} 
    \usepackage[dvipsnames, table]{xcolor}  

    \def\paperID{16095}
    \def\confName{CVPR}
    \def\confYear{2024}
}{}
\iftoggle{eccv}{
    \usepackage[review,year=2024,ID=9588]{eccv}
    \usepackage{booktabs}
}{}
\iftoggle{corl}{
    \usepackage[dvipsnames,table]{xcolor}
    \definecolor{mydarkblue}{rgb}{0,0.08,0.45}
    \usepackage{corl_2024} 
}{}

\usepackage{multirow}  
\usepackage{amsmath} 
\usepackage{amssymb}  
\usepackage{acro}  
\usepackage{nicefrac}  
\usepackage[percent]{overpic}  
\usepackage{pict2e}  
\usepackage{tikz}  
\usetikzlibrary{spy, decorations.text}
\usepackage{booktabs}
\iftoggle{cvpr}{
    \usepackage{stfloats}
}{}

\iftoggle{cvpr}{
    \definecolor{cvprblue}{rgb}{0.21,0.49,0.74}
    \usepackage[pagebackref,breaklinks,colorlinks,citecolor=cvprblue]{hyperref}
}{}
\iftoggle{eccv}{
    \usepackage[pagebackref,breaklinks,colorlinks,citecolor=eccvblue]{hyperref}
    \usepackage{orcidlink}  
}{}

\DeclareAcronym{SfM}{
  short=SfM,
  long=Structure from Motion,
}

\DeclareAcronym{SLAM}{
  short=SLAM,
  long=Simultaneous Localization and Mapping,
}

\DeclareAcronym{VIO}{
  short=VIO,
  long=Visual Inertial Odometry,
}

\DeclareAcronym{DoG}{
  short=DoG,
  long=Difference of Gaussians,
}
\DeclareAcronym{FAST}{
  short=FAST,
  long=Features from Accelerated Segment Test,
}
\DeclareAcronym{ORB}{
  short=ORB,
  long=Oriented FAST and Rotated BRIEF,
}
\DeclareAcronym{SIFT}{
  short=SIFT,
  long=Scale Invariant Feature Transform,
}

\DeclareAcronym{BRISK}{
  short=BRISK,
  long=Binary Robust Invariant Scalable Keypoints,
}

\DeclareAcronym{NRE}{
  short=NRE,
  long=Neural Reprojection Error,
}
\DeclareAcronym{MLP}{
  short=MLP,
  long=Multi-Layer Perceptron,
}
\DeclareAcronym{GNN}{
  short=GNN,
  long=Graph Neural Network,
}

\DeclareAcronym{CNN}{
  short=CNN,
  long=Convolutional Neural Network,
}

\DeclareAcronym{NN}{
  short=NN,
  long=nearest neighbor,
}

\DeclareAcronym{NLL}{
  short=NLL,
  long=Negative Log Likelihood,
}

\DeclareAcronym{AUC}{
    short=AUC,
    long=area under the cumulative error curve,
}

\DeclareAcronym{ATE}{
    short=ATE,
    long=absolute trajectory error,
}

\DeclareAcronym{SVD}{
    short=SVD,
    long=singular value decomposition,
}

\DeclareAcronym{TUM}{
    short=TUM,
    long=Technical University Munich,
}

\DeclareAcronym{MOTS}{
    short=MOTS,
    long=Multi Object Tracking and Segmentation,
}

\def\frameworkname{DynamicGlue}
\iftoggle{cvpr}{
    \title{DynamicGlue: Epipolar and Time-Informed Data Association \\in Dynamic Environments using Graph Neural Networks}
}{}
\iftoggle{eccv}{
    \title{DynamicGlue - Epipolar and Time-Informed Data Association in Dynamic Environments using Graph Neural Networks}
}{}
\iftoggle{corl}{
    \title{DynamicGlue - Epipolar and Time-Informed Data Association in Dynamic Environments using Graph Neural Networks}
}{}

\author{Theresa Huber$^*$\\
{\tt\small theresa.huber@tum.de}
\and
Simon Schaefer$^*$\\
{\tt\small simon.k.schaefer@tum.de}
\and 
Stefan Leutenegger \\
{\tt\small stefan.leutenegger@tum.de} 
}

\begin{document}
\maketitle

\begin{abstract}
The assumption of a static environment is common in many geometric computer vision tasks like SLAM but limits their applicability in highly dynamic scenes. Since these tasks rely on identifying point correspondences between input images within the static part of the environment, we propose a graph neural network-based sparse feature matching network designed to perform robust matching under challenging conditions while excluding keypoints on moving objects. We employ a similar scheme of attentional aggregation over graph edges to enhance keypoint representations as state-of-the-art feature-matching networks but augment the graph with epipolar and temporal information and vastly reduce the number of graph edges. 
Furthermore, we introduce a self-supervised training scheme to extract pseudo labels for image pairs in dynamic environments from exclusively unprocessed visual-inertial data. A series of experiments show the superior performance of our network as it excludes keypoints on moving objects compared to state-of-the-art feature matching networks while still achieving similar results regarding conventional matching metrics. When integrated into a SLAM system, our network significantly improves performance, especially in highly dynamic scenes.
\end{abstract}

\iftoggle{cvpr}{\vspace{-0.7cm}}{}
\iftoggle{corl}{}{}    
\section{Introduction}
\label{sec:introduction}
In the realm of geometric computer vision, establishing correspondences between image keypoints serves as a foundational element for various tasks, including \ac{SLAM} and \ac{SfM}. This data association process enables the inference of relative transformations between images of a moving camera and the underlying environmental structure. Despite significant strides in the field, challenges persist, such as handling large viewpoint changes, occlusion, weak texture, and dynamic objects within the scene.

\iftoggle{cvpr}{

\begin{figure}
    \centering
    \begin{tikzpicture}[spy using outlines={circle,black,connect spies}]
        \node {\includegraphics[width=0.48\textwidth]{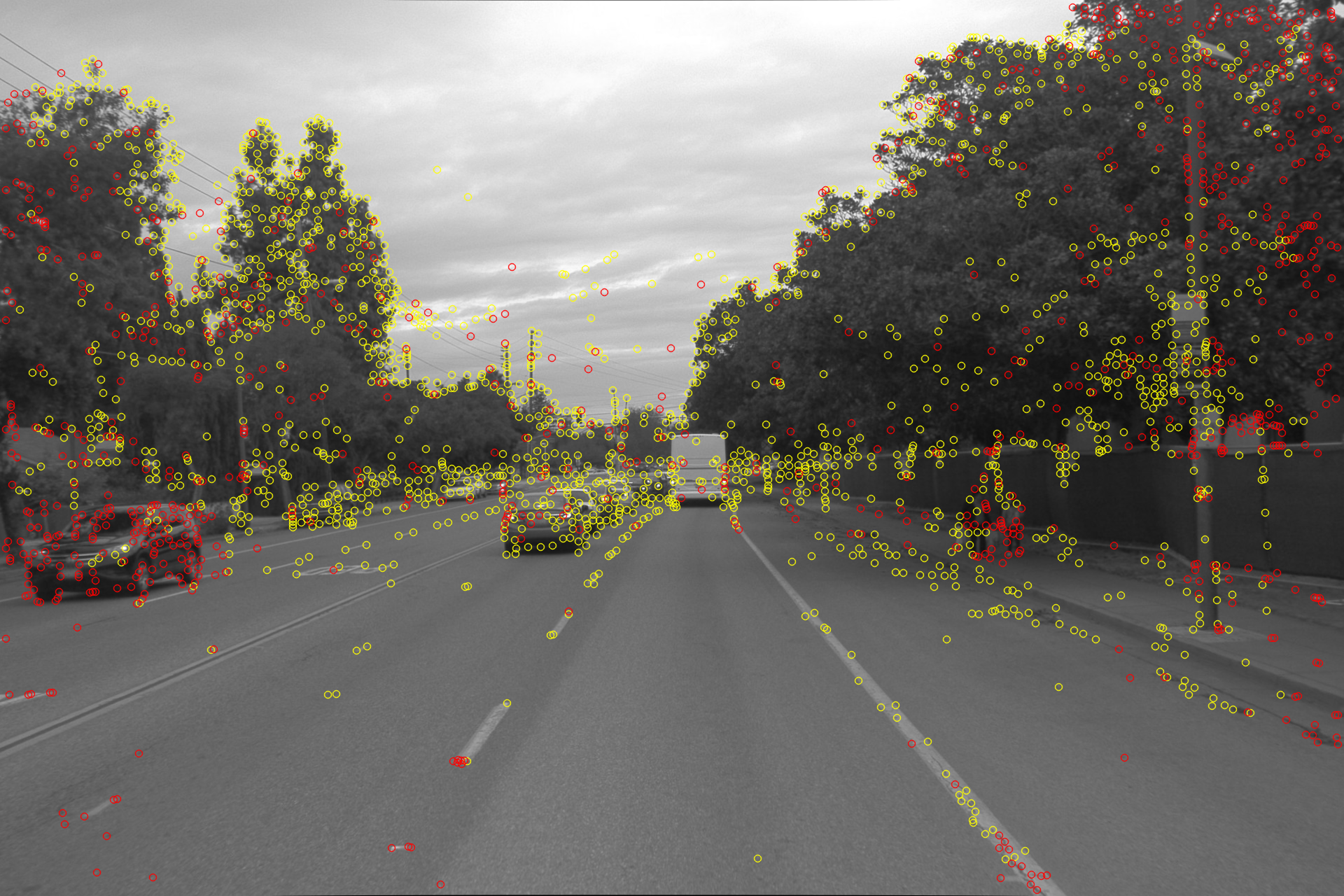}};

        \spy[size=2cm, magnification=1.5] on (-3.4,-0.7) in node [left] at (-0.5,1.3);
        \node at (-1.5, 2.5) {\color{BrickRed}\textbf{Dynamic}};

        \spy[size=2cm, magnification=1.5] on (-0.75,-0.4) in node [left] at (2,-1.3);
        \node at (1, -2.5) {\color{Yellow}\textbf{Static}};
    \end{tikzpicture}
    \caption{In this paper, we present \frameworkname, a matching framework for dynamic scenes. Compared to state-of-the-art approaches, our framework cannot only deal with large changes in appearance, such as viewpoint changes but also differentiate dynamic from static parts of the scene. Matched keypoints are shown in yellow, and unmatched keypoints in red.}
    \label{fig:qualitative}
    \vspace{-0.5cm}
\end{figure}
}{}

\iftoggle{eccv}{
\begin{figure}
    \centering
    \begin{tikzpicture}[spy using outlines={circle,black,connect spies}]
        \node {\includegraphics[width=0.88\textwidth]{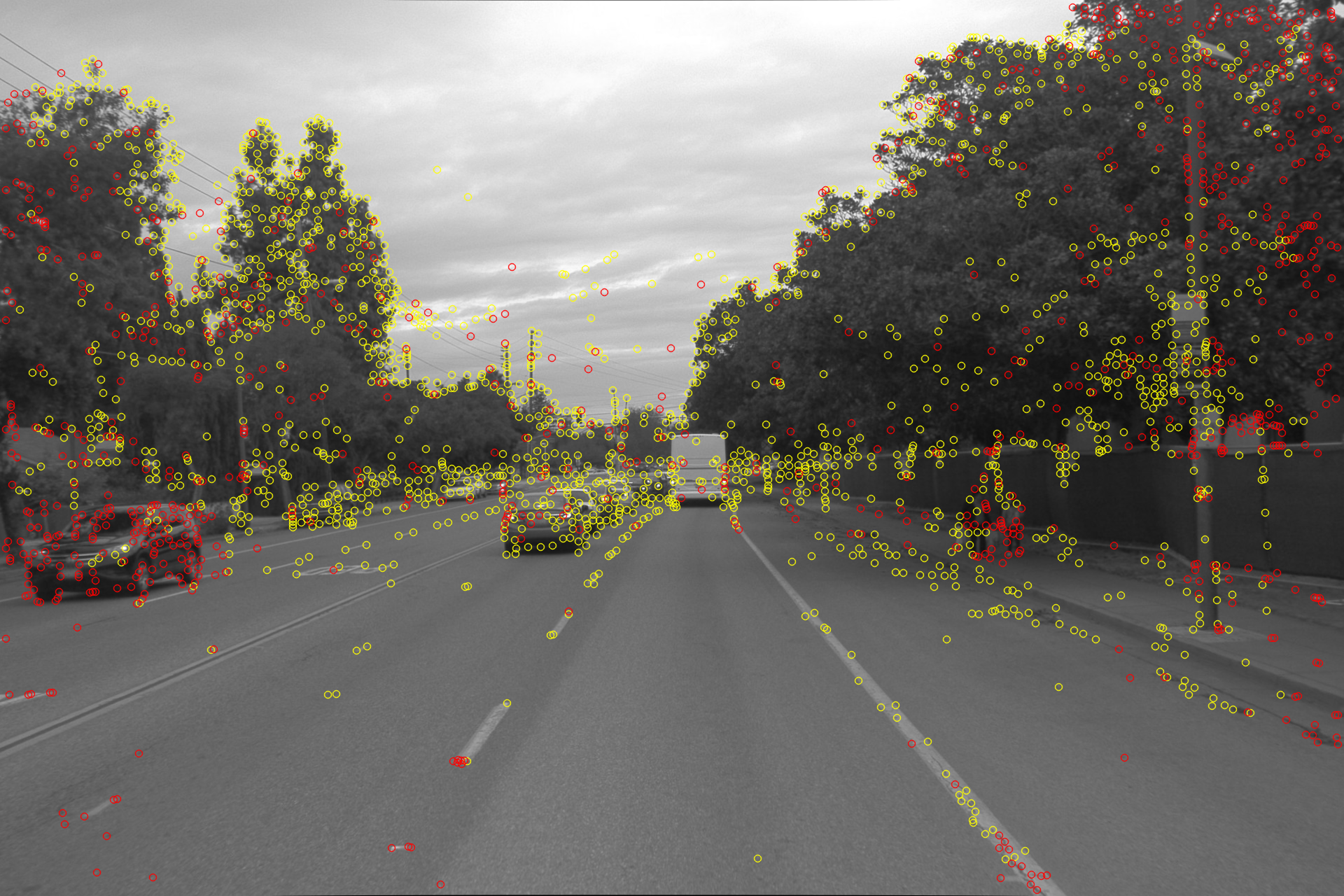}};

        \spy[size=2cm, magnification=1.5] on (-4.4,-0.9) in node [left] at (-0.5,1.3);
        \node at (-1.5, 2.5) {\color{BrickRed}\textbf{Dynamic}};

        \spy[size=2cm, magnification=1.5] on (-0.9,-0.4) in node [left] at (2,-1.3);
        \node at (1, -2.5) {\color{Yellow}\textbf{Static}};
    \end{tikzpicture}
    \caption{In this paper, we present \frameworkname, a matching framework for dynamic scenes. Compared to state-of-the-art approaches, our framework cannot only deal with large changes in appearance, such as viewpoint changes but also differentiate dynamic from static parts of the scene, e.g., moving cars on the opposing lane from cars waiting for a traffic light. Matched keypoints are shown in yellow, unmatched keypoints in red.}
    \label{fig:qualitative}
    \vspace{-0.5cm}
\end{figure}
}{}

\iftoggle{corl}{
\begin{figure}
    \centering
    \begin{tikzpicture}[spy using outlines={circle,black,connect spies}]
        \node {\includegraphics[trim={0 6cm 0 8cm},clip,width=0.88\textwidth]{media/matching_qualitative/waymo/ours/5072733804607719382_5807_570_5827_570/465.png}};

        \spy[size=2cm, magnification=1.0] on (-5,-0.9) in node [left] at (-1.5,1.3);
        \node at (-1.5, 2.5) {\color{BrickRed}\textbf{Dynamic}};

        \spy[size=2cm, magnification=1.5] on (-1.0,-0.7) in node [left] at (2,-1.3);
        \node at (1, -2.5) {\color{Yellow}\textbf{Static}};
    \end{tikzpicture}
    \caption{In this paper, we present \frameworkname, a matching framework for dynamic scenes. Compared to state-of-the-art approaches, our framework cannot only deal with large changes in appearance, such as viewpoint changes, but also differentiate dynamic from static parts of the scene, e.g., moving cars on the opposing lane from cars waiting for a traffic light. Matched keypoints are shown in yellow, unmatched keypoints in red.}
    \label{fig:qualitative}
    \vspace{-0.5cm}
\end{figure}

}{}


While existing methodologies have addressed these challenges to some extent, a notable limitation is their tendency to assume a static environment. This assumption poses a hurdle for downstream tasks like \ac{SLAM} and \ac{VIO}, which heavily rely on identifying correspondences within the static portion of the environment. Common approaches either employ RANSAC to filter out dynamic elements as outliers or resort to masking predefined semantic classes, such as vehicles, potentially sacrificing valuable information due to misclassification due to no differentiation between e.g.\ parking and moving cars and lack of generalization. 

In response, we present a novel context-aware keypoint matching framework designed to not only navigate substantial changes in appearance but also distinguish between static and dynamic elements in the scene. Drawing inspiration from SuperGlue~\citep{Sarlin2020}, we construct a graph from the keypoints of two input images. We enhance their representations through self- and cross-attentional aggregation, additionally incorporating temporal and epipolar geometry consistency information as edge features within a \ac{GNN}.

Furthermore, we introduce a self-supervised pipeline for labeling point correspondences on image pairs within dynamic environments, leveraging raw stereo-inertial sensor data to eliminate the need for human annotation. 
We summarize our contributions as follows:
\iftoggle{corl}{
\textbf{(1)} We propose a graph neural network-based sparse feature matching network architecture. While inspired by SuperGlue~\cite{Sarlin2020}, we augment the graph to include epipolar and temporal information, allowing us to vastly decrease the size of the graph and number of graph operations, making it much more computationally efficient and aware of the scene dynamics.
\textbf{(2)} We introduce a self-supervised training scheme that facilitates training using exclusively unprocessed real-world images and IMU data. To achieve this, we create pseudo labels by leveraging a SLAM system in combination with off-the-shelf networks for depth prediction and multi-object tracking.
\textbf{(3)} We showcase the superior performance of our framework across a wide range of scenarios, including highly dynamic environments. While our system exhibits comparable performance to state-of-the-art methods when assessed using conventional matching metrics for a static world, it significantly outperforms them by reducing the number of potentially misleading matches on moving objects by $65\%$. Furthermore, we integrate our approach into a SLAM system, highlighting its ability to enhance its overall performance (VIO by $29\%$ and VI-SLAM by $15\%$).
}{
\begin{itemize}
    \item We propose a graph neural network-based sparse feature matching network architecture. While inspired by SuperGlue~\cite{Sarlin2020}, we augment the graph to include epipolar and temporal information, allowing us to vastly decrease the size of the graph and number of graph operations, making it much more computationally efficient and aware of the scene dynamics.
    \item We introduce a self-supervised training scheme that facilitates training using exclusively unprocessed real-world images and IMU data. To achieve this, we create pseudo labels by leveraging a SLAM system in combination with off-the-shelf networks for depth prediction and multi-object tracking.
    \item We showcase the superior performance of our framework across a wide range of scenarios, including highly dynamic environments. While our system exhibits comparable performance to state-of-the-art methods when assessed using conventional matching metrics for a static world, it significantly outperforms them by reducing the number of potentially misleading matches on moving objects by $65\%$. Furthermore, we integrate our approach into a SLAM system, highlighting its ability to enhance its overall performance (VIO by $29\%$ and VI-SLAM by $15\%$).
\end{itemize}
}
\section{Related Work}
\label{sec:related_work}
\iftoggle{cvpr}{\begin{figure*}[t!]
    \includegraphics[width=\textwidth]{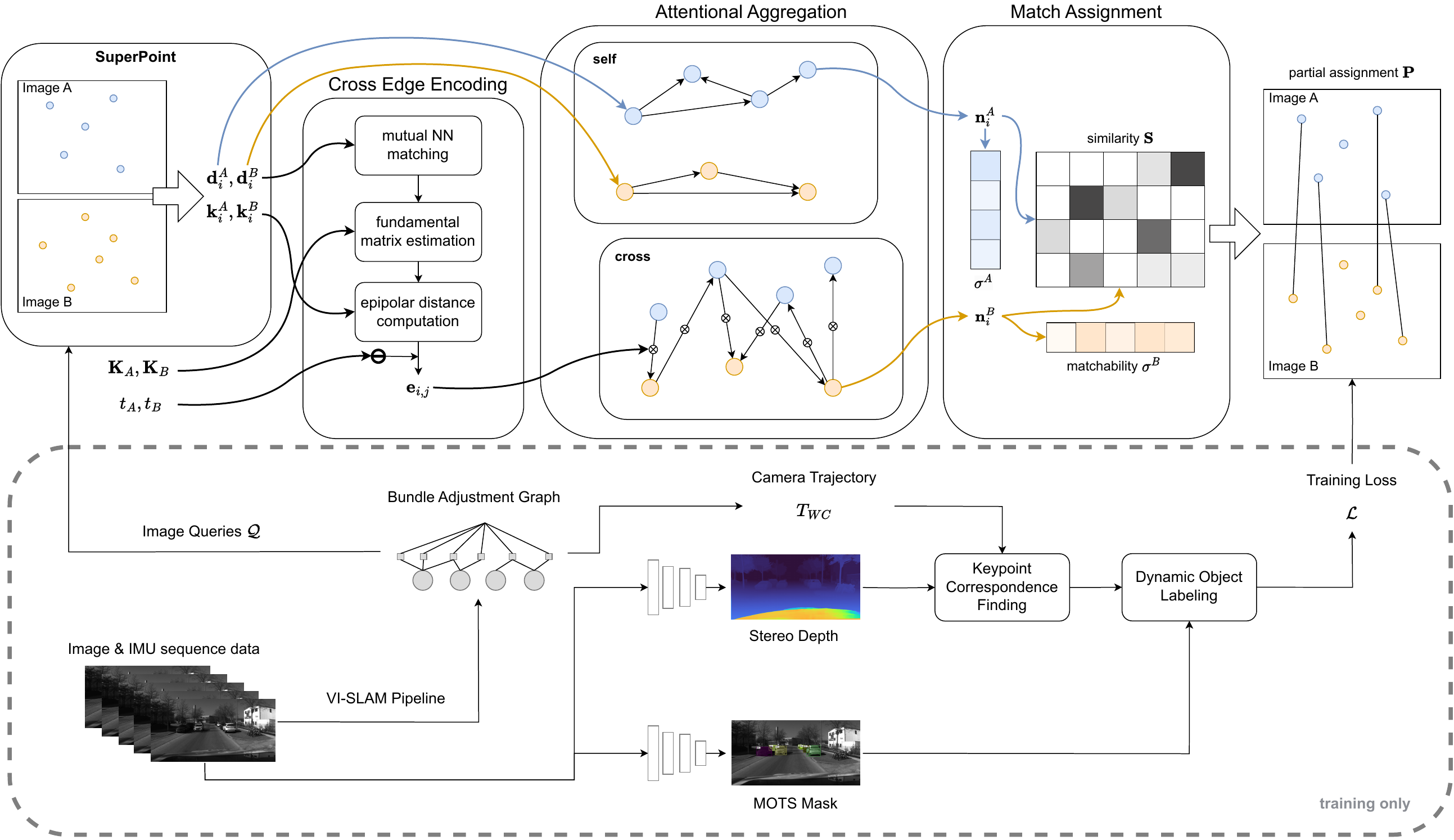}
    \caption{The network architecture comprises three parts: Graph formation, attentional aggregation, and match assignment. 
    After creating the graph from Superpoint~\cite{DeTone2018} keypoints, epipolar and temporal information in the first step, an enhanced representation $\mathbf{n}$ of the initial descriptors $\mathbf{d}$ is computed using attentional aggregation over self and cross-edges in the second step. The third part computes a partial assignment based on the enhanced keypoint encodings.}
    \label{fig:architecture}
    \iftoggle{corl}{\vspace{-0.5cm}}{}
\end{figure*}}{}
Many data association pipelines follow a common scheme to detect salient keypoints, compute feature descriptors, and then match the feature descriptors between images, often followed by an outlier rejection step.
Based on the found correspondences, a relative transformation is estimated by using, for example, the 8-point algorithm.
Regarding the estimation of scene dynamics from 2D images, optical flow-based methods have been proposed \cite{Zachary2020, Ilg2017, Xu2022}. 
However, to estimate the dynamics of the current image, they require a sequence of subsequent images and, therefore, introduce an additional delay to the system.
While traditionally handcrafted methods for feature description and matching have been used \cite{Harris1988, Mouats2018, Lowe2004, Leutenegger2011}, more and more parts of this pipeline have applied deep learning to increase performance in recent years. 
In SuperPoint~\cite{DeTone2018}, a deep feature descriptor is learned with a multi-step scheme, first using synthetically generated data of basic shapes with groundtruth keypoint detections, then on random homographies of real-world images. 
Similarly, NRE~\cite{Germain2021} learns a deep descriptor end-to-end in conjunction with a relative camera pose estimation. 
While several methods for matching directly based on feature descriptors have been proposed, \cite{Lowe2004, Sarlin2020}, they fail to include contextual information extracted from image pairs.
Yi et al.~\cite{Yi2018a} are the first to include contextual information in the matching problem, such as the camera motion and scene geometry, and learn to predict the essential matrix using direct supervision.
While this improves the matching performance, they do not encode scene dynamics information, arguably the most relevant information in many downstream tasks.
Patch2Pix~\cite{Zhou2021} and NRE~\cite{Germain2021} both follow an end-to-end approach. Instead of learning correspondences, they directly regress the relative pose between two images.
Since one main drawback of these end-to-end correspondence networks is a low pixel accuracy in the matching due to memory and runtime limitations, they propose a two-step network in a detect-to-refine manner.
While this approach is promising for complex and highly dynamic environments, it still requires weak supervision. It can hardly be integrated into optimization-based downstream applications, such as \ac{SLAM}, that further refine the camera trajectory based on global optimization.
In \cite{Schmidt2017}, Schmidt et al.\ were the first to propose a fully self-supervised training pipeline for data association based on a dense \ac{SLAM} pipeline, similar to ours.
However, they assumed a static environment, requiring RGB-D sensor input, which is generally more scarce, and using a much more computationally complex dense \ac{SLAM} pipeline.
Recently, Superglue~\cite{Sarlin2020} and its successors \cite{Lindenberger2023, Liu2020, Liu2020_2} have dominated the field by leveraging \ac{GNN}s for matching that both include local as well as global contextual information, in the form of self- and cross-edges. 
Our work, in contrast, combines the node-level information with geometric and temporal edge encodings to explicitly create awareness of the scene geometry and dynamics. 
Furthermore, we do not require any direct supervision or synthetically generated data for training.
Thereby, we achieve good generalization capabilities as well as scalability by training directly on real-world data and are much more computationally lightweight.


\section{Problem Formulation}
\label{sec:problem_formulation}
\iftoggle{eccv}{\begin{figure*}[t!]
    \includegraphics[width=\textwidth]{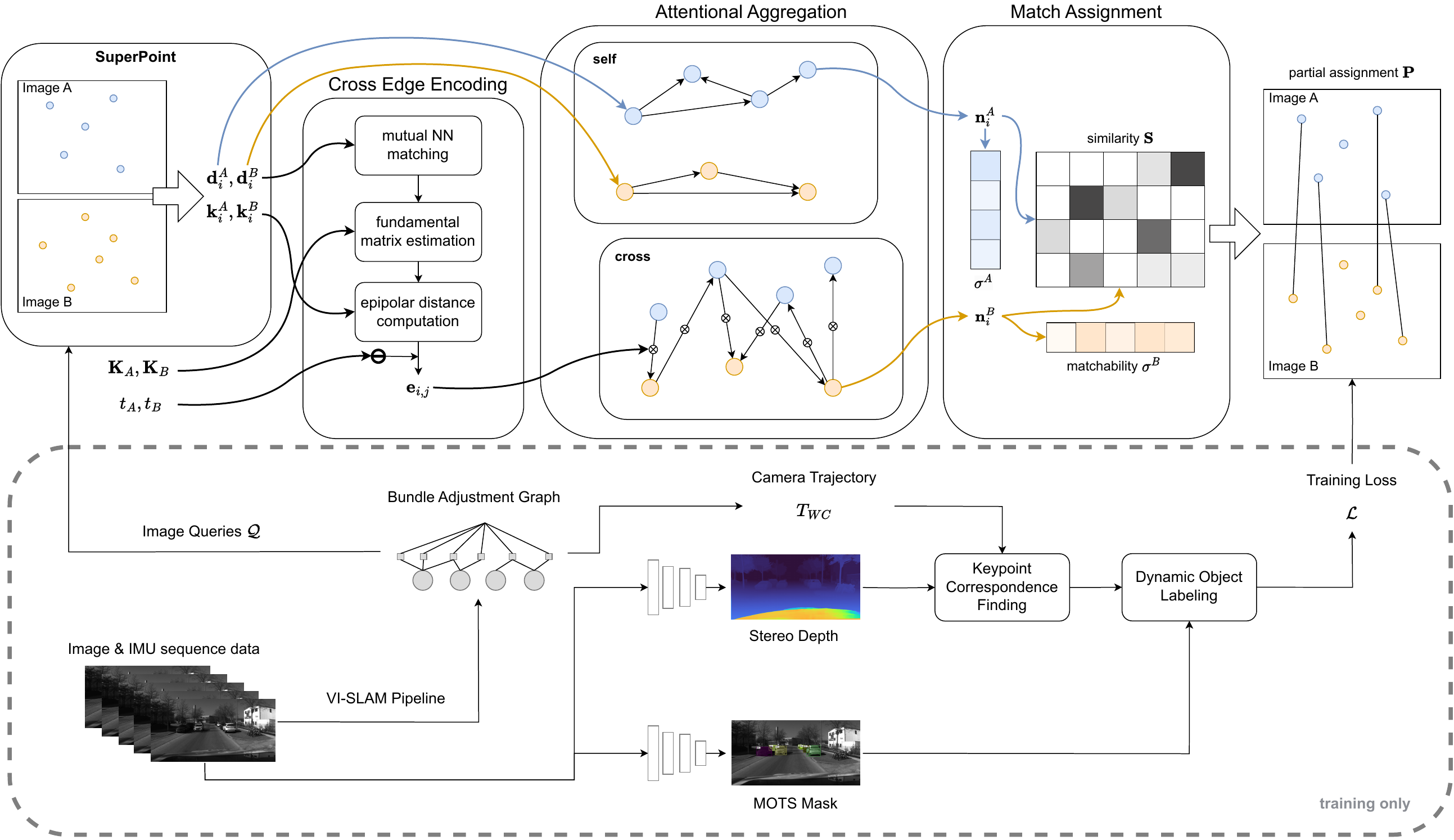}
    \caption{The network architecture comprises three parts: Graph formation, attentional aggregation, and match assignment. 
    After creating the graph from Superpoint~\cite{DeTone2018} keypoints, epipolar and temporal information in the first step, an enhanced representation $\mathbf{n}$ of the initial descriptors $\mathbf{d}$ is computed using attentional aggregation over self and cross-edges in the second step. The third part computes a partial assignment based on the enhanced keypoint encodings.}
    \label{fig:architecture}
    \iftoggle{corl}{\vspace{-0.5cm}}{}
\end{figure*}}{}
\iftoggle{corl}{\begin{figure*}[t!]
    \includegraphics[width=\textwidth]{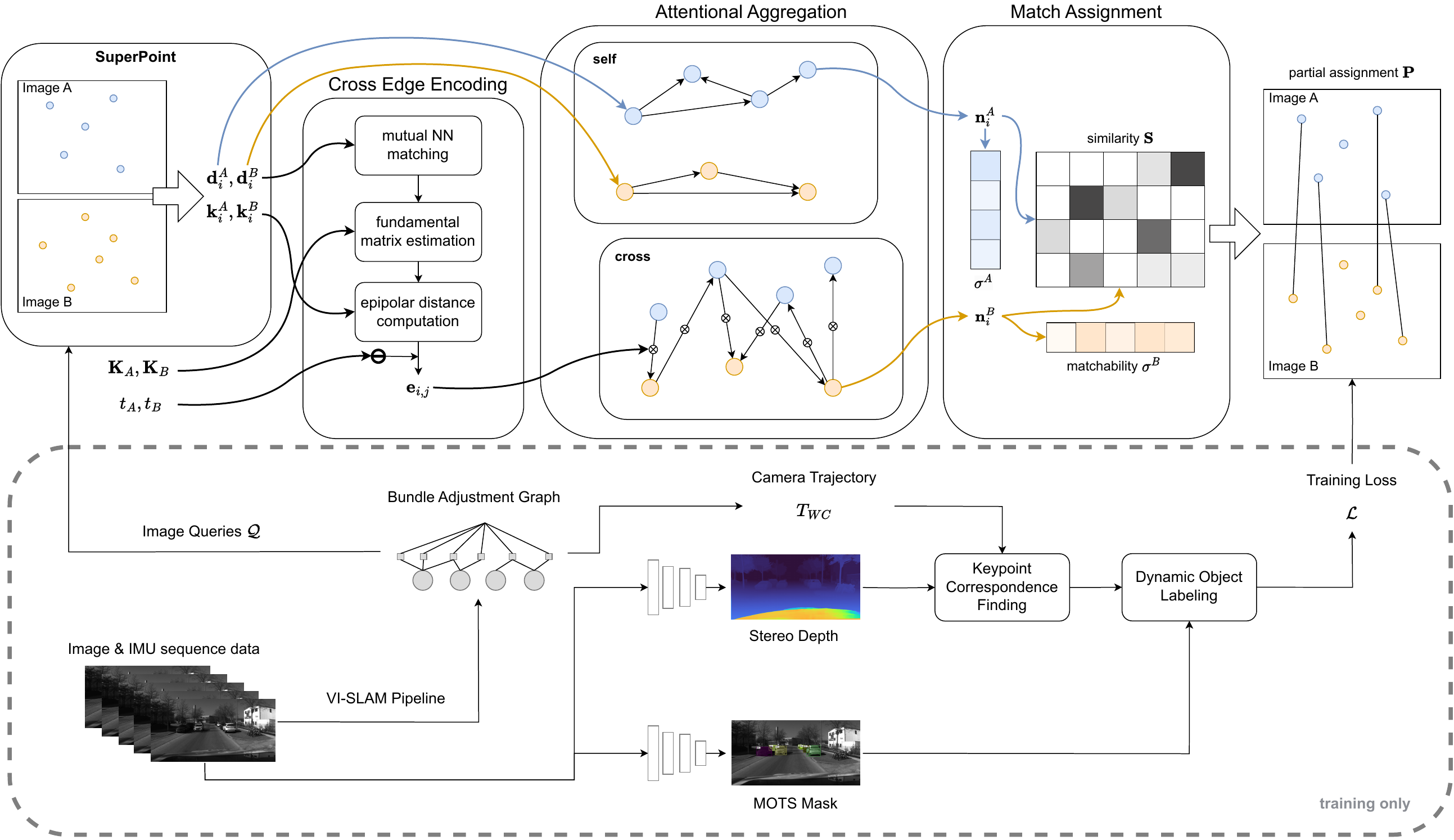}
    \caption{The network architecture comprises three parts: Graph formation, attentional aggregation, and match assignment. 
    After creating the graph from Superpoint~\cite{DeTone2018} keypoints, epipolar and temporal information in the first step, an enhanced representation $\mathbf{n}$ of the initial descriptors $\mathbf{d}$ is computed using attentional aggregation over self and cross-edges in the second step. The third part computes a partial assignment based on the enhanced keypoint encodings.}
    \label{fig:architecture}
    \iftoggle{corl}{\vspace{-0.5cm}}{}
\end{figure*}}{}
The data association problem is finding a partial assignment between the $N$ keypoints of image $\mathcal{I}_A$ and $M$ keypoints of image $\mathcal{I}_B$.
The keypoints are indexed by $\mathcal{A}:=\{1, \cdots, N\} $ and $\mathcal{B}:=\{N+1, \cdots, N+M\} $, respectively, and defined by their position $k_i \, \forall i \in \mathcal{A}\cup \mathcal{B}$ and the SuperPoint~\cite{DeTone2018} descriptor $d_i \, \forall i \in \mathcal{A}\cup \mathcal{B}$. 
The camera intrinsics $\mathbf{K}_A$ and $\mathbf{K}_B$ and the image timestamps $t_A$ and $t_B$ are known.
Each keypoint in one image can be matched to, at most, one keypoint in the other image since a correct match results from the projection of a unique 3D location. 
It may also be matched to no keypoint due to occlusions or since it was not detected in the other image, resulting in a set of matches $\mathcal{M} = \{(i,j)\} \subset \mathcal{A} \times \mathcal{B}$. 
To this end, the network estimates a partial assignment matrix $\mathbf{P} \in [0,1]^{N\times M}$ between keypoints of images $\mathcal{I}_A$ and $\mathcal{I}_B$ like in SuperGlue~\cite{Sarlin2020}.

\section{Epipolar and Time-Informed GNN}
\label{sec:eti_gnn}
Figure \ref{fig:architecture} shows our network architecture, which consists of three stages. 
The graph formation builds a graph using the keypoints and their descriptors, epipolar and temporal information. 
This graph is then processed using attentional aggregation to extract high-dimensional node embeddings for each keypoint.
These node embeddings are then matched to each other across the two images in the last stage, leading to the final associations.

\subsection{Graph Formation}
We create a graph with nodes defined as the SuperPoint~\cite{DeTone2018} keypoints detected in the input images and two types of directed edges. 
Self-edges $\mathcal{E}_\text{self}$ which connect keypoints within the same image, and cross-edges $\mathcal{E}_\text{cross}$ which connect keypoints in one image to keypoints in the other.
Apart from SuperGlue~\cite{Sarlin2020} or LightGlue~\cite{Lindenberger2023}, we do not create a complete graph but create cross-edges only to the $10$ most similar keypoints in the other image based on the SuperPoint~\cite{DeTone2018} descriptors and self-edges only to the $10$ closest keypoints in the same image based on the Euclidean distance between keypoint positions in the image.
We incorporate information about the consistency with the epipolar geometry and time information in the \ac{GNN} as edge features for cross-edges. 
To this end, a lightweight matching is performed between the keypoints of both images, resulting in a set of matches $\mathcal{M}$, where each match is assigned a weight $w_m \;\forall m \in \mathcal{M}$ based on descriptor similarity.
Based on the found matches $\mathcal{M}$ and match-weights $w_m$, the fundamental matrix $F$ is computed using the weighted 8-point algorithm.
Using this estimate of the fundamental matrix $\mathbf{F}$, for every cross-edge $(i,j) \in \mathcal{E}_\text{cross}$, the symmetric epipolar distance $d_\text{epi}(\mathbf{F}, k_i, k_j)$ is computed between the keypoints $k_i$ and $k_j$ which are represented by node $i$ and $j$ respectively.
For numerical stability, we apply the logarithm on the computed epipolar distance.
\begin{equation}
    d_\text{epi}(\mathbf{F}, k_i, k_j) = d(k_i, \mathbf{F}, k_j)^2 + d(k_j, \mathbf{F}^T, k_i)^2 \text{.}
\end{equation}
Even for inaccurate initial estimates of the fundamental matrix, we found that including the epipolar information as edge features provides valuable information to cluster the match candidates in inliers and outliers.
However, to further quantify the accuracy of the initial estimation, the sum of all weights $w_m$ used for the 8-point algorithm and the mean and minimum of epipolar distances are added to the edge features. 
Furthermore, the passed time between the two images is incorporated as the difference of timestamps $t_i$ and $t_j$.
The edge features $\mathbf{e}_{i,j}$ are defined by concatenating all aforementioned information:
\begin{equation}
    \mathbf{e}_{i,j} = \begin{bmatrix} 
    \log(d_\text{epi}(\mathbf{F}, k_i, k_j)) \\
    \frac{1}{ \mid\mathcal{E}_\text{cross} \mid} \sum_{(a,b) \in \mathcal{E}_\text{cross}} \log(d_\text{epi}(\mathbf{F}, k_a, k_b))  \\ 
    \min_{(a,b) \in \mathcal{E}_\text{cross}} \log(d_\text{epi}(\mathbf{F}, k_a, k_b)) \\
    \sum_{w_m \in \mathcal{M}} w_m  \\
    t_j - t_i 
    \end{bmatrix}
    \text{.}
\end{equation}

\subsection{Attentional Aggregation}
The node embeddings are initialized as the SuperPoint~\cite{DeTone2018} descriptors and alternately updated by attentional aggregation over the self- and cross-edges, similar to the residual message passing proposed in SuperGlue~\cite{Sarlin2020}. 
For the aggregation of self-edges, we use the graph transformer introduced in \cite{Shi2020} with multi-head attention weights $\alpha_{i,j}$: 
\iftoggle{corl}{
\begin{align}
    \mathbf{\hat{m}}_{\mathcal{E}_\text{self}, i} = \mathbf{W}_1 \mathbf{n_i} + \sum_{j\in \mathcal{E}_\text{self}(i)} \alpha_{i,j}^\text{s} \mathbf{W}_2 \mathbf{n_j}, \hskip0.1\textwidth
    \alpha_{i,j}^\text{s} = \text{sm} \left( \frac{(\mathbf{W}_3 \mathbf{n_i})^T(\mathbf{W}_4 \mathbf{n_j})}{\sqrt{D}} \right) \text{,}
\end{align}
}{
\begin{align}
    \mathbf{\hat{m}}_{\mathcal{E}_\text{self}, i} &= \mathbf{W}_1 \mathbf{n_i} + \sum_{j\in \mathcal{E}_\text{self}(i)} \alpha_{i,j}^\text{s} \mathbf{W}_2 \mathbf{n_j} \\
    \alpha_{i,j}^\text{s} &= \text{sm} \left( \frac{(\mathbf{W}_3 \mathbf{n_i})^T(\mathbf{W}_4 \mathbf{n_j})}{\sqrt{D}} \right) \text{,}
\end{align}
}
where $D$ is the dimensionality of the node embeddings $\mathbf{n}$, \text{sm} the softmax activation function and $\mathbf{W}_i$ learnable network parameters.
\iftoggle{corl}{
For cross-edge aggregation, we additionally attend to edge features.
}{
We alter the graph transformer for cross-edge aggregation to attend to edge features, too. 
\begin{align}
    \mathbf{\hat{m}}_{\mathcal{E}_\text{cross}, i} &= \mathbf{W}_5 \mathbf{n_i} + \sum_{j\in \mathcal{E}_\text{cross}(i)} \alpha_{i,j}^\text{c} \left( \mathbf{W}_6 \mathbf{n_j} + \mathbf{W}_7 \mathbf{e}_{i,j} \right) \\
    \alpha_{i,j}^\text{c} &= \text{sm} \left( \frac{(\mathbf{W}_8 \mathbf{n_i})^T(\mathbf{W}_9 \mathbf{n_j} + \mathbf{W}_{10} \mathbf{e}_{i,j})}{\sqrt{D}} \right) \text{.}
\end{align}
}
The graph transformer layer is followed by a \textit{PairNorm}~\cite{Zhao2019} layer to prevent over-smoothing of node embeddings.

\subsection{Match Assignment Head}
We use the lightweight assignment head introduced by LightGlue~\cite{Lindenberger2023} to estimate correspondences between the keypoints of the two images based on the updated node embeddings $\mathbf{n}$. 
This head computes the partial assignment $\mathbf{P}$ by predicting a matchability score $\sigma$ for every keypoint and similarity score matrix $\mathbf{S} \in \mathbb{R}^{N \times M}$ between all pairs of keypoints. 
The elements in $\mathbf{S}$ encode how likely a pair of points correspond, whereas the matchability score $\sigma$ encodes how likely a keypoint has a correspondence in the other image.
For all $i \in \mathcal{A}$, $j \in \mathcal{B}$, and $o \in \mathcal{A} \cup \mathcal{B}$ we have:
\begin{align}
    \mathbf{S}_{ij} &= \text{Linear}(\textbf{n}_i)^T \text{Linear}(\textbf{n}_j) \\
    \sigma_o &= \text{Sigmoid}(\text{Linear}(\textbf{n}_o))\text{,}
\end{align}
where $\text{Linear}(\cdot)$ denotes a learnable linear transformation with bias.
A low matchability score,  $\sigma_i \rightarrow 0$, is expected when a keypoint is occluded or undetected in the other image.
The combination of similarity score $\mathbf{S}$ and matchability score $\sigma$ results in the partial assignment matrix $\mathbf{P}$:
\begin{equation}
    \mathbf{P}_{ij} = \sigma_i \sigma_j \underset{l \in \mathcal{A}}{\text{sm}}(\mathbf{S}_{lj})_i \underset{l \in \mathcal{B}}{\text{sm}}(\mathbf{S}_{il})_j \text{.}
\end{equation}
To extract correspondences, all pairs $(i,j)$ that mutually yield the highest assignment score $\mathbf{P}_{ij}$ within their row and column are selected if $\mathbf{P}_{ij}$ is larger than a threshold $\tau$.

\section{Self-Supervised Training}
\label{sec:training}
We propose a self-supervised training scheme using datasets containing merely sequences of real-world stereo images and IMU data. 
As outlined in Figure \ref{fig:architecture}, the datasets are pre-processed to extract pairs of images, denoted as image queries $\mathcal{Q}$, used for training the network. 
Furthermore, a set of labels containing groundtruth matches $\mathcal{M}_{\text{gt}}$ and sets of non-matchable keypoints $\mathcal{N}_{\text{gt}}^A$ and $\mathcal{N}_{\text{gt}}^B$ in each image respectively are created from the input data. 

\subsection{SLAM-based Pseudo-Groundtruth Generation}
Modern VI-\ac{SLAM} systems can achieve centimeter-level accuracy after offline bundle adjustment, even for long sequences.
Therefore, we suggest using the bundle-adjusted factor graph of OKVIS2~\cite{Leutenegger2022} for pseudo groundtruth generation.
Each node in the graph represents a state within the sequence and is defined by its timestamp and optimized 6D pose. 
Each edge contains edges representing the reprojection errors, where keypoints in the images are assigned to landmarks in the reconstruction. 
For training data generation, image queries are extracted by parsing the bundle adjustment graph of a session and selecting any pair of camera images that track common landmarks.
To contain the number of extracted samples to a tractable amount and decrease the similarity of image queries, the number of common landmarks must be above a threshold $c$, and only every $s$-th state is considered for the second image of a query.
Also, the bundle adjustment graph defines the keypoint locations in each image as the projection of the tracked landmarks back into the camera image.

We found that the number of groundtruth matches from landmark projections is typically too small for good generalization capabilities due to the relatively sparse detections and matches from OKVIS2~\cite{Leutenegger2022} for long sequences.
Thus, we augment them with SuperPoint~\cite{DeTone2018} keypoints.
Each keypoint from image $\mathcal{I}_A$ is projected into image $\mathcal{I}_B$ using the estimated depth map, created from the stereo depth network IGEV-Stereo~\cite{xu2023iterative}, from the timestamp of image $\mathcal{I}_A$ and the relative transformation extracted from the bundle adjustment graph.
If a keypoint in image $\mathcal{I}_B$ coincides with the projected keypoint, the pair is labeled as a groundtruth match. 
If the point is projected outside of image $\mathcal{I}_B$ or there is no keypoint in a radius of 50 pixels in image $\mathcal{I}_B$, this point is labeled as non-matchable. 
All other keypoints remain unlabelled. 
To prevent outliers in the groundtruth due to inaccuracies in the depth maps, the same is performed for all keypoints in image $\mathcal{I}_B$ with the according depth map. 
Only classifications verified by both projections are considered valid.

\subsection{Groundtruth in a Dynamic Environment}
The network shall be trained to not match keypoints on moving objects since these matches will be outliers for the static world assumption used by a \ac{SLAM} system.
For queries with only a short time difference, which are typical for \ac{SLAM} systems processing subsequent images, the movement of the dynamic objects may be so small that the projective approach may also label matches on the moving objects as correct. 
The moving objects are explicitly extracted to prevent this, and their keypoints are excluded from the groundtruth matches. 
To this end, the \ac{MOTS} network STEm-Seg~\cite{Athar_Mahadevan20ECCV} computes instance segmentation masks for objects within the image sequence.
Due to limitations of the \ac{MOTS} model, this work focuses on cars and pedestrians only but can easily be extended to other semantic classes.
3D point clouds are extracted using the created depth maps based on the segmentation masks assigned to the processed instance. 
These are then back-projected to point clouds in a common coordinate frame to compensate for the camera movement between the frames, using the optimized inter-frame transformation in the bundle-adjusted graph. 
To identify the moving instances, the distance between all combinations of extracted point clouds is computed with the Chamfer distance $d_{\text{pcl}}$. 
The object is labeled \emph{moving} if $d_{\text{pcl}}$ is larger than a threshold of $5$ meters.
For all image queries for which the time difference is not zero, all keypoints lying on a moving object are excluded from $\mathcal{M}_{\text{gt}}$ and added to the set of non-matchable keypoints $\mathcal{N}_{\text{gt}}$.
For image queries with no time difference, there are no inter-frame movements so that all keypoints in $\mathcal{M}_{\text{gt}}$ are considered matchable.

\subsection{Loss Formulation}
Based on the generated labels $\mathcal{M}_{\text{gt}}$, $\mathcal{N}_{\text{gt}}^A$, and $\mathcal{N}_{\text{gt}}^B$ the \ac{NLL} loss is minimized during training acting on the predicted partial assignment score $\mathbf{P}$ and matchability scores $\sigma$ for images $\mathcal{I}_A$ and $\mathcal{I}_B$
\iftoggle{corl}{
, with the overall loss being the sum $\mathcal{L} = -(\mathcal{L}_{\text{M}} + \mathcal{L}_{\text{N}}^A + \mathcal{L}_{\text{N}}^B)$.
\begin{align}
    \mathcal{L}_{\text{M}} = \left(\frac{1}{| \mathcal{M}_{\text{gt}} | } \sum_{(i,j) \in \mathcal{M}_{\text{gt}}} \log \mathbf{P}_{i,j} \right), \hskip0.05\textwidth
    \mathcal{L}_{\text{N}} = \left(\frac{1}{2| \mathcal{N}_{\text{gt}} | } \sum_{i \in \mathcal{N}_{\text{gt}}} \log (1 - \sigma_i) \right)
    \label{eq:loss}
\end{align}
}{
.
\begin{align}
    \mathcal{L}_{\text{M}} &= \left(\frac{1}{| \mathcal{M}_{\text{gt}} | } \sum_{(i,j) \in \mathcal{M}_{\text{gt}}} \log \mathbf{P}_{i,j} \right) \\
    \mathcal{L}_{\text{N}} &= \left(\frac{1}{2| \mathcal{N}_{\text{gt}} | } \sum_{i \in \mathcal{N}_{\text{gt}}} \log (1 - \sigma_i) \right) \\
    \mathcal{L} &= -(\mathcal{L}_{\text{M}} + \mathcal{L}_{\text{N}}^A + \mathcal{L}_{\text{N}}^B)
    \label{eq:loss}
\end{align}
}

\section{Experiments}
\label{sec:experiments}
We implement our network using the PyG library~\cite{Fey2019} in the Torch framework~\cite{Paszke_PyTorch_An_Imperative_2019}.
The network is trained using Adam~\cite{adam_opt} with a learning rate of $0.0001$ and a batch size of $32$.

\iftoggle{corl}{
We train our model on the respective training set from TUM4Seasons dataset~\cite{wenzel2020fourseasons} and the Hilti-Oxford dataset~\cite{hilti2022} due to their versatility, as they provide stereo-inertial sensor data and contain significant appearance changes.
For evaluation, we additionally use the Waymo Open Perception~\cite{Sun_2020_CVPR} as well as our own in-the-wild dataset to test the generalization performance of our approach. 
We refer to the appendix for a detailed description of all datasets.
}{
We trained the model on the respective training set from the TUM4Seasons dataset~\cite{wenzel2020fourseasons} and the Hilti-Oxford dataset~\cite{hilti2022} due to their versatility, as they provide stereo-inertial sensor data and contain significant appearance changes.
While the TUM4Seasons dataset~\cite{wenzel2020fourseasons} contains scenes recorded from a car's perspective on different roads in various conditions, the Hilti-Oxford dataset~\cite{hilti2022} is recorded from a handheld device showing indoor and outdoor scenes on construction sites and buildings. 
We set the thresholds to $c=10$ for both datasets and $s=37$ for the TUM4Seasons dataset~\cite{wenzel2020fourseasons} and $s=13$ for the Hilti-Oxford dataset~\cite{hilti2022}. 
While Hilti-Oxford contains more rapid viewpoint changes due to the handheld camera, TUM4Seasons was recorded on a car, such that subsequent images are more similar.
Using the proposed training scheme, we generate roughly $185,000$ image pairs from the TUM4Seasons dataset and $50,000$ image pairs from the Hilti-Oxford dataset in the training set.
Furthermore, we extract the sequences with ego-motion from the Waymo Open Perception~\cite{Sun_2020_CVPR} dataset as an additional dataset for evaluation that was not used during training. 
This dataset shows inner and outer-city scenes from a car perspective in the USA and provides extensive labels for evaluating matches on dynamic objects.
Similarly, we have recorded our own in-the-wild dataset to further test the generalization performance of our network with regard to diverse dynamics object types. 
}

\subsection{Matching Performance}
\begin{table*}
    \footnotesize
    \centering
    \def\matchingCellColor{\cellcolor{gray!10}}
    \begin{tabular}{lccccccc}
        \toprule
        & AUC@5\textdegree $\uparrow$ & AUC@10\textdegree $\uparrow$ & AUC@20\textdegree $\uparrow$& P $\uparrow$ & MS $\uparrow$ & M\textsubscript{mov} $\downarrow$ & K\textsubscript{mov} $\downarrow$ \\ \midrule
        \multicolumn{8}{c}{\matchingCellColor \textbf{Hilti-Oxford Dataset~\cite{hilti2022}}} \\ \midrule
Mutual NN & 29.62 & 41.58 & 51.81 & 72.73 & 24.33 & 0.84 & 17.53	\\
LightGlue~\cite{Lindenberger2023} & \textbf{34.98} & \textbf{47.94} & \textbf{58.54} & 62.75 & \textbf{35.72} &  1.15 & 41.81 \\ \midrule  
\frameworkname & 33.41 & 46.45 & 57.24 & \textbf{77.38} & 27.08 & \textbf{0.62} & \textbf{12.89}
 \\ \midrule
        \multicolumn{8}{c}{\matchingCellColor \textbf{TUM4Seasons Dataset~\cite{wenzel2020fourseasons}}} \\ \midrule
Mutual NN 	 & 47.78 & 58.75 & 67.45 & 79.38 & 20.48 & 3.77 & 36.78 \\
LightGlue~\cite{Lindenberger2023} & \textbf{78.84} & \textbf{87.60} & \textbf{92.69} & 88.88 & \textbf{40.17} & 2.68 & 47.11 \\ \midrule 
DG - No edge features & 70.14 & 80.43 & 87.31 & 93.61 & 21.55 & 0.96 & 10.67 \\
DG - No time & 69.66 & 79.97 & 86.95 & 93.14 & 22.11 & 1.04 & 11.78 \\
DG - No pair norm & 69.18 & 79.52 & 86.52 & \textbf{94.74} & 20.56 & 0.84 & 9.26 \\
DG - iterative & 70.28 & 80.58 & 87.39 & 94.36 & 22.62 & 1.15 & 12.15 \\ \midrule
\frameworkname & 69.70 & 79.98 & 86.90 & 93.87 & 21.22 & \textbf{0.59} & \textbf{6.97}
 \\ \midrule
        \multicolumn{8}{c}{\matchingCellColor \textbf{Waymo Open Perception Dataset~\cite{Sun_2020_CVPR}}} \\ \midrule
Mutual NN & 44.05 & 52.47 & 59.02 & 62.03 & 13.83 & \textbf{6.95} & 14.03 \\
LightGlue~\cite{Lindenberger2023} & 43.18 & \textbf{57.87} & \textbf{69.23} & \textbf{85.09} & \textbf{47.47} & 8.36 & 47.15 \\ \midrule  
\frameworkname & \textbf{45.17} & 53.30 & 59.49 & 74.84 & 13.45 & 7.48 & \textbf{11.69}
 \\ \midrule
        \multicolumn{8}{c}{\matchingCellColor \textbf{Self-Recorded In-The-Wild Dataset}} \\ \midrule
Mutual NN & 20.47 & 34.67 & 48.59 & 68.74 & 17.99 & 1.93 & 10.66 \\
LightGlue~\cite{Lindenberger2023} & \textbf{26.73} & \textbf{44.05} & \textbf{58.85} & 65.18 & \textbf{33.98} & 3.35 & 41.37 \\ \midrule 
\frameworkname & 26.43 & 42.16 & 56.55 & \textbf{76.84} & 21.06 & \textbf{1.02} & \textbf{7.18} 
 \\
        \bottomrule
    \end{tabular}
    \caption{Evaluation of matching performance. The best results are marked in bold. While achieving state-of-the-art results in the commonly used matching metrics, our method outperforms the baselines in regards to the metrics that take into account the scene dynamics, $M\textsubscript{mov}$ and $K\textsubscript{mov}$.}
    \label{tab:eval_matching}
    \vspace{-0.5cm}
\end{table*}
We evaluate the network's performance based on the precision, the matching score, and the \ac{AUC} of pose errors, as commonly used in the area.
While the precision $\text{P}$ defines the ratio of correct matches to found matches, the matching score $\text{MS}$ relates it to the number of keypoints in image $\mathcal{I}_A$.
Thereby, a match is classified as correct if its epipolar distance is below $5\cdot 10^{-4}$. 
We estimate the essential matrix with RANSAC based on the computed correspondences and select the maximum errors in the rotational and translational components to compute the pose error, represented as an angular error.
\nottoggle{corl}{The \ac{AUC} of pose errors is reported at three thresholds: 5\textdegree, 10\textdegree, and 20\textdegree.}

Additionally, we evaluate the matching in dynamic environments based on the matches of keypoints on moving objects. 
To this end, we propose two new metrics:
\iftoggle{corl}{
\begin{align}
    \text{M}_\text{mov} &=  \frac{|\mathcal{M}_\text{mov}|}{|\mathcal{M}|} \text{,} &\text{K}_\text{mov} = \frac{|\mathcal{A}_\text{mov}| + |\mathcal{B}_\text{mov}|}{|\mathcal{A}| + |\mathcal{B}|} \text{,}
\end{align}
}{
\begin{align}
    \text{M}_\text{mov} &=  \frac{|\mathcal{M}_\text{mov}|}{|\mathcal{M}|} \text{,} \\
    \text{K}_\text{mov} &= \frac{|\mathcal{A}_\text{mov}| + |\mathcal{B}_\text{mov}|}{|\mathcal{A}| + |\mathcal{B}|} \text{,}
\end{align}
}
with $|\mathcal{M}|$ the number of predicted matches, $|\mathcal{A}|$ and $|\mathcal{B}|$ the number of keypoints in $\mathcal{I}_A$ and $\mathcal{I}_B$, respectively, and $|\mathcal{A}_\text{mov}|$ and $|\mathcal{B}_\text{mov}|$ the keypoints on moving objects. 
These metrics allow us to evaluate the ability of the matcher to exclude moving objects from the matching.

For a fair comparison to state-of-the-art frameworks for feature matching, including naive descriptor matching based on mutual nearest neighbor, all of the tested methods use the same descriptor SuperPoint~\cite{DeTone2018}.
In addition, we have abducted several ablation studies. 
We evaluate the effect of omitting the edge features, the temporal information, and the pair norm. 
In addition, we test an alternative network structure in which the edge features are iteratively updated after each layer based on re-evaluating the nearest neighbors of each node.

LightGlue~\cite{Lindenberger2023} is trained on more diverse and much larger datasets (about 5 million image pairs). 
Therefore, for a fair comparison, we re-train LightGlue~\cite{Lindenberger2023} on our dataset while maintaining their architecture and loss formulation.
For an image query with $n$ keypoints per image, it computes $4n^2$ edges to connect to every keypoint in the same image and to every keypoint in the other image. 
This results in 4,000,000 edges for a typical image query with 1,000 keypoints, compared to 20,000 edges in our method. 
While being much more computationally efficient, our method significantly outperforms state-of-the-art feature-matching networks with regards to matches on dynamic objects as indicated by $\text{D}_\text{mov}$ and $\text{K}_\text{mov}$, resulting in an improvement of $85\%$ to learned approaches and $47\%$ to \ac{NN} matching. 
While LightGlue~\cite{Lindenberger2023} is optimized to return as many matches as possible, resulting in high AUC scores, our method yields more trustworthy matches, which leads to much higher precision across most datasets.
\iftoggle{corl}{
Qualitative examples of our method's performance in regards to differentiating static from dynamic objects can be found in the appendix.
}{
More qualitative examples of the performance of our method in regards to differentiating static from dynamic objects can be found in Figure \ref{fig:qualitative_examples}.
}

\iftoggle{eccv}{
\subsection{Runtime Comparison}
\label{sec:runtime}
The runtimes are the compute time required to match keypoints in a pair of frames $\mathcal{I}_\text{A}$ and $\mathcal{I}_\text{B}$ for a varying number of keypoints, evaluated on an NVIDIA RTX A4000 several times and using 10 runs as a warmup.
LightGlue~\cite{Lindenberger2023} already shows significant improvements over SuperGlue~\cite{Sarlin2020}, mostly due to early stopping and pruning. 
However, due to the fully connected internal graph structure, their computational complexity increases quadratically with the number of keypoints.  
In contrast, our method uses a sparsely connected graph and consequently scales only linearly with the number of keypoints in the images.
Thus, we achieve up to $4$ times smaller runtime compared to LightGlue~\cite{Lindenberger2023} and up to $30$ smaller runtime compared to SuperGlue~\cite{Sarlin2020}, see Figure \ref{fig:okvis_trajectory}, enabling real-time processing even on small onboard processors of robots.
}{}

\subsection{SLAM Integration}
\iftoggle{cvpr}{
\iftoggle{cvpr}{

\begin{table*}[bp]
    \footnotesize
    \centering
    \begin{tabular}{lcl|cccc|cccc}
    \multirow{2}{*}{\textbf{Session}} & \multirow{2}{*}{\textbf{Label}} & & \multicolumn{4}{c|}{\textbf{VIO}} & \multicolumn{4}{c}{\textbf{SLAM}} \\
     & & & \textbf{Okvis2 [11]} & \textbf{LG [13]} & \textbf{MNN} & \textbf{DG (Ours)} & \textbf{Okvis2 [11]} & \textbf{LG [13]} & \textbf{MNN} & \textbf{DG (Ours)} \\ \midrule
    2021-05-10\_19-15-19 & l & & \textbf{1.39} & 1.69 & 1.66 & 1.75 & \textbf{1.02} & 1.07 & 1.08 & 1.15 \\
    2021-01-07\_13-12-23 & l & & 11.04 & 10.23 & 11.05 & \textbf{7.71} & \textbf{1.37} & 3.01 & 3.14 & 2.49 \\ 
    2020-04-07\_11-33-45 & m & & 61.81 & 111.54 & 106.39 & \textbf{37.64} & 76.71 & 90.83 & 85.15 & \textbf{33.67} \\
    2020-06-12\_10-10-57 & m & & 29.01 & 14.63 & \textbf{12.24} & 13.34 & 4.98 & 13.52 & \textbf{4.45} & 6.20 \\
    2021-05-10\_18-32-32 & m & & 5.42 & 4.90 & 5.28 & \textbf{4.86} & 5.12 & 5.09 & \textbf{4.67} & 5.05 \\
    2021-01-07\_10-49-45 & h & & 34.82 & 30.95 & 21.59 & \textbf{10.77} & 14.00 & \textbf{10.20} & 10.38 & 12.42 
    \end{tabular}

    \caption{Absolute trajectory errors for various sessions of TUM4Seasons [23] using Okvis2 [11] with several different matching frameworks: LG (LightGlue [13]), MNN (Mutual NN), and DG (\frameworkname). Our method clearly outperforms the baseline matchers, especially in the VIO case and for medium to highly dynamic scenes.}
    \label{tab:okvis_ate_all}
\end{table*}
}{}

\iftoggle{eccv}{
\begin{table*}[bp]
    \footnotesize
    \centering
    \begin{tabular}{lcl|cccc|cccc}
    \toprule
    \multirow{2}{*}{\textbf{Session}} & \multirow{2}{*}{\textbf{Label}} & & \multicolumn{4}{c|}{\textbf{VIO}} & \multicolumn{4}{c}{\textbf{SLAM}} \\
     & & & \textbf{OK2} & \textbf{LG} & \textbf{MNN} & \textbf{DG} & \textbf{OK2} & \textbf{LG} & \textbf{MNN} & \textbf{DG} \\ \midrule
     2021-05-10/19-15-19 & l & & \textbf{1.4} & 1.7 & 1.6 & 1.7 & \textbf{1.0} & 1.1 & 1.1 & 1.1 \\
     2021-01-07/13-12-23 & l & & 11.0 & 10.2 & 11.0 & \textbf{7.7} & \textbf{1.4} & 3.0 & 3.1 & 2.5 \\ 
     2020-04-07/11-33-45 & m & & 61.8 & 111.5 & 106.4 & \textbf{37.6} & 76.7 & 90.8 & 85.2 & \textbf{33.7} \\
     2020-06-12/10-10-57 & m & & 29.0 & 14.6 & \textbf{12.2} & 13.3 & 5.0 & 13.5 & \textbf{4.4} & 6.2 \\
     2021-05-10/18-32-32 & m & & 5.4 & \textbf{4.9} & 5.3 & \textbf{4.9} & 5.1 & 5.1 & \textbf{4.7} & 5.0 \\
     2021-01-07/10-49-45 & h & & 34.8 & 30.9 & 21.6 & \textbf{10.8} & 14.0 & \textbf{10.2} & 10.4 & 12.4 \\ \bottomrule
    \end{tabular}
    
    \caption{Absolute trajectory errors for various sessions of TUM4Seasons \cite{wenzel2020fourseasons} using OK2 (Okvis 2 \cite{Leutenegger2022}) with several different matching frameworks: LG (LightGlue \cite{Lindenberger2023}), MNN (Mutual NN), and DG (\frameworkname). Our method clearly outperforms the baseline matchers, especially in the VIO case and for medium to highly dynamic scenes.}
    \label{tab:okvis_ate_all}
\end{table*}
}{}

\iftoggle{corl}{
\begin{figure}
\begin{minipage}{\textwidth}
\begin{minipage}{0.29\textwidth}
    \centering
    \includegraphics[width=0.9\textwidth]{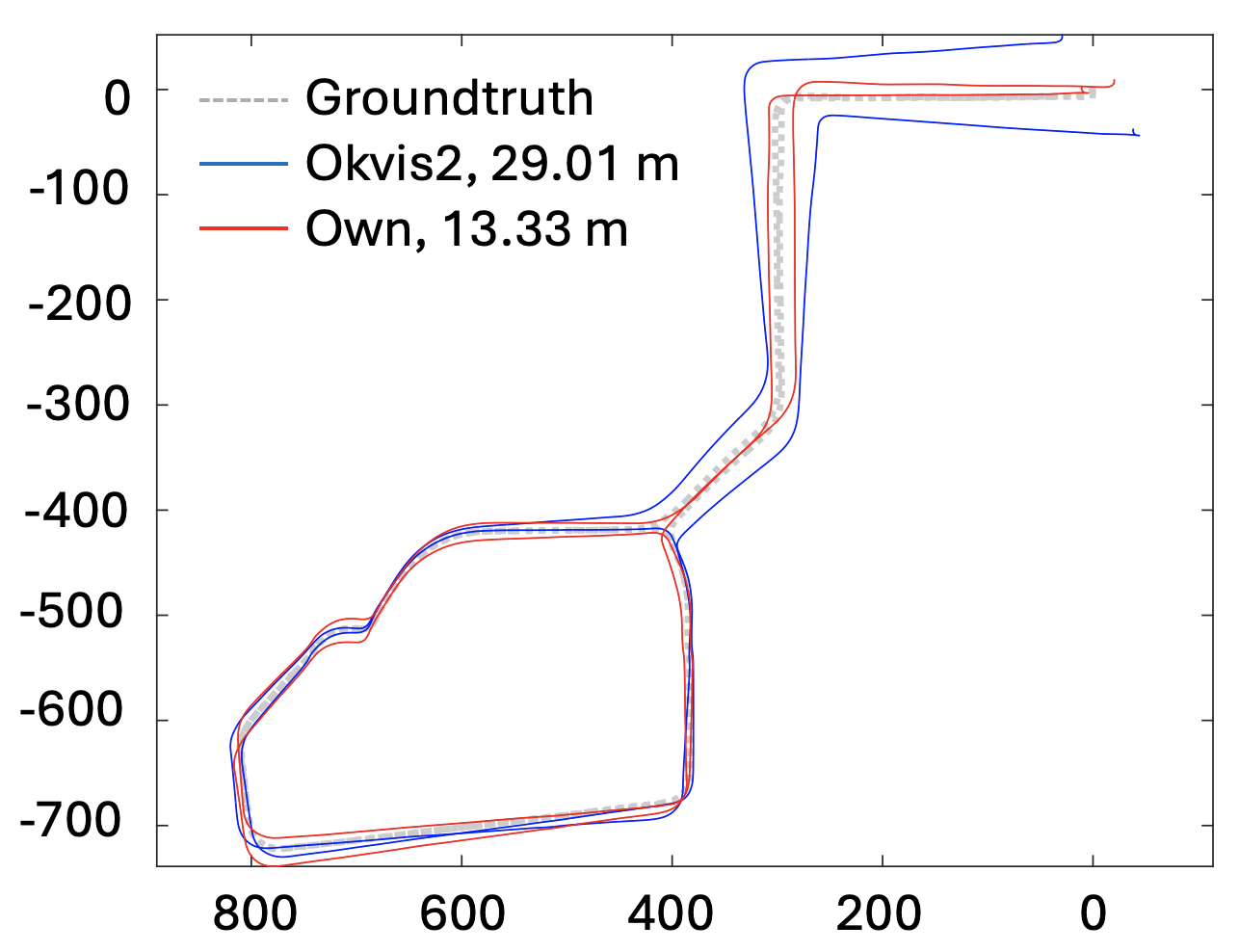}
\end{minipage}
\hfill
\begin{minipage}{0.70\textwidth}
    \centering
    \tiny
    \begin{tabular}{lcl|cccc|cccc}
    \toprule
    \multirow{2}{*}{\textbf{Session}} & \multirow{2}{*}{\textbf{Label}} & & \multicolumn{4}{c|}{\textbf{VIO}} & \multicolumn{4}{c}{\textbf{SLAM}} \\
     & & & \textbf{OK2} & \textbf{LG} & \textbf{MNN} & \textbf{DG} & \textbf{OK2} & \textbf{LG} & \textbf{MNN} & \textbf{DG} \\ \midrule
     19-15-19 & l & & \textbf{1.4} & 1.7 & 1.6 & 1.7 & \textbf{1.0} & 1.1 & 1.1 & 1.1 \\
     13-12-23 & l & & 11.0 & 10.2 & 11.0 & \textbf{7.7} & \textbf{1.4} & 3.0 & 3.1 & 2.5 \\ 
     11-33-45 & m & & 61.8 & 111.5 & 106.4 & \textbf{37.6} & 76.7 & 90.8 & 85.2 & \textbf{33.7} \\
     10-10-57 & m & & 29.0 & 14.6 & \textbf{12.2} & 13.3 & 5.0 & 13.5 & \textbf{4.4} & 6.2 \\
     18-32-32 & m & & 5.4 & \textbf{4.9} & 5.3 & \textbf{4.9} & 5.1 & 5.1 & \textbf{4.7} & 5.0 \\
     10-49-45 & h & & 34.8 & 30.9 & 21.6 & \textbf{10.8} & 14.0 & \textbf{10.2} & 10.4 & 12.4 \\ \bottomrule
    \end{tabular}
\end{minipage}
\end{minipage}
\caption{\textbf{Left:} Trajectories of our method vs. the baseline OKVIS2~\cite{Leutenegger2022}. Our method drifts significantly less than the baseline, reducing especially the azimuth error. \textbf{Right:} Absolute trajectory errors for various sessions of TUM4Seasons \cite{wenzel2020fourseasons} using OK2 (Okvis 2 \cite{Leutenegger2022}) with several matching frameworks: LG (LightGlue \cite{Lindenberger2023}), MNN (Mutual NN), and DG (\frameworkname). Our method clearly outperforms the baselines, especially in the VIO case and high dynamic scenes.}
\label{fig:okvis_trajectory}
\label{tab:okvis_ate_all}
\vspace{-0.5cm}
\end{figure}
}{}
\begin{figure}
    \centering
    \iftoggle{cvpr}{
        \includegraphics[width=0.32\textwidth]{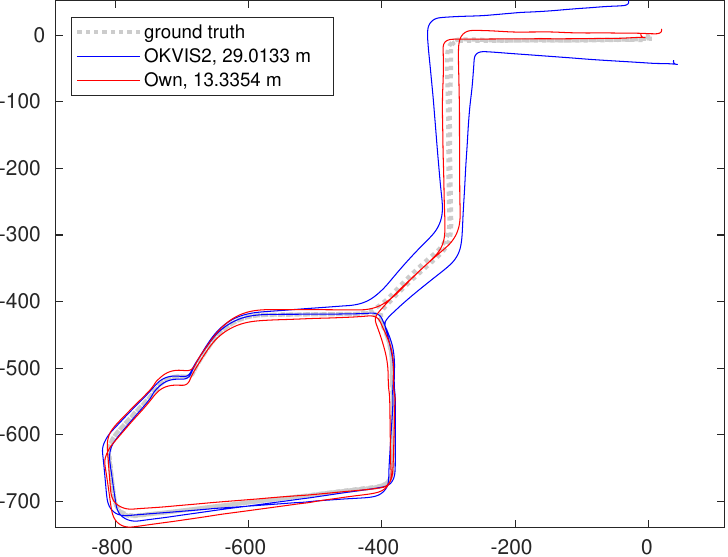}
        \includegraphics[width=0.32\textwidth]{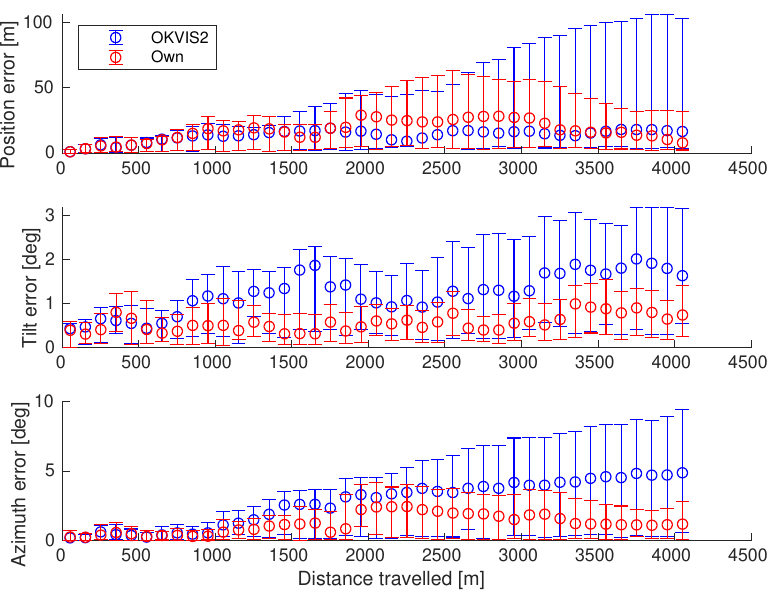}
        \caption{Trajectories and drift statistics of our method vs. the baseline OKVIS2~\cite{Leutenegger2022} on the test sequence  `2020-06-12\_10-10-57` of the TUM4Season dataset~\cite{wenzel2020fourseasons}. The relative motion errors for position and orientation are aggregated over different sub-trajectory lengths. Our method drifts significantly less than the baseline, reducing especially the azimuth error.}
    }{}
    \iftoggle{eccv}{
        \includegraphics[width=0.32\textwidth]{media/recording_2020-06-12_10-10-57_traj.pdf}
        \includegraphics[width=0.32\textwidth]{media/recording_2020-06-12_10-10-57_error_plots.pdf}
        \includegraphics[width=0.32\textwidth]{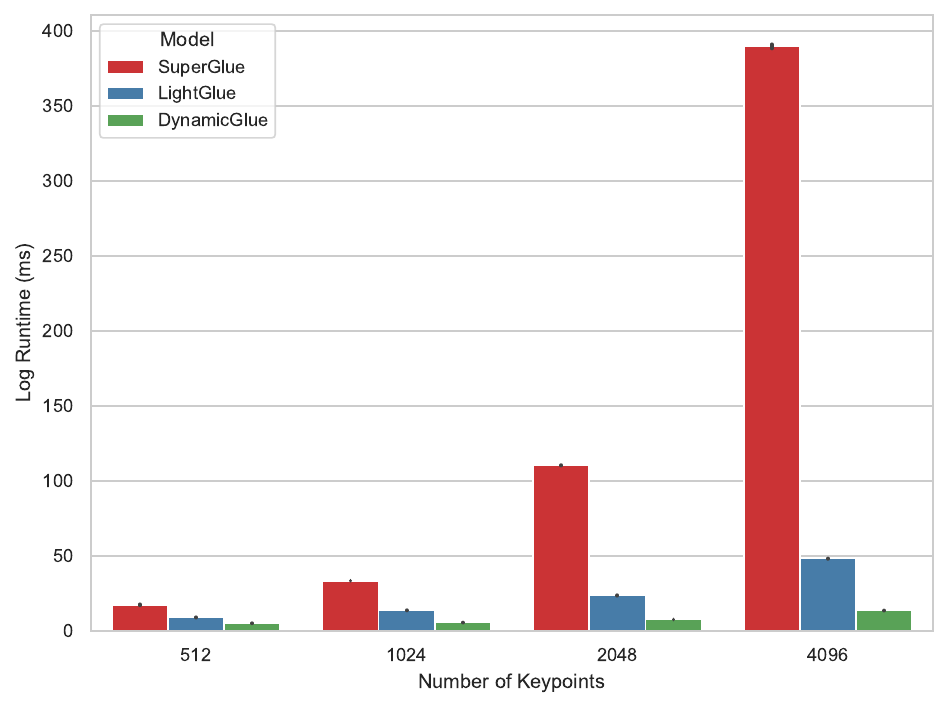}
        \caption{Left: Trajectories and drift statistics of our method vs. the baseline OKVIS2~\cite{Leutenegger2022} on the test sequence  `2020-06-12\_10-10-57` of the TUM4Season dataset~\cite{wenzel2020fourseasons}. The relative motion errors for position and orientation are aggregated over different sub-trajectory lengths. Our method drifts significantly less than the baseline, reducing especially the azimuth error. Right: Runtime comparison of our framework in comparison to Superglue~\cite{Sarlin2020} and LightGlue~\cite{Lindenberger2023}.}
    }{}
    \label{fig:okvis_trajectory}
    \vspace{-0.5cm}
\end{figure}
}{}
To demonstrate the effectiveness of our matching framework in real-world applications, we have integrated our pipeline as well as the baselines into the state-of-the-art SLAM system OKVIS2~\cite{Leutenegger2022, zhang_2023_hilti_challenge}. 
OKVIS2 originally uses BRISK~\cite{Leutenegger2011} keypoints that are matched by computing the Hamming distance between descriptors. 
For the integration, BRISK is replaced by SuperPoint~\cite{DeTone2018} and then matched using the integrated matching framework.
When matching an image to the map, the matching procedure had to be adapted to account for the additional contextual information used in our framework.
Landmarks are not handled individually but clustered by frames so the network matches complete images.

The results are provided for sessions of the TUM4Seasons dataset from the validation and test split, comparing the original OKVIS2 with BRISK as the baseline and the adapted version with our network.
The sessions are processed in \ac{VIO} mode with loop closures disabled and in \ac{SLAM} mode with loop closures enabled. 
We provide relative trajectory error statistics in Figure \ref{fig:okvis_trajectory} that quantify the odometry drift that is especially meaningful when processing without loop closures.
Furthermore, the \ac{ATE} is reported in Table \ref{tab:okvis_ate_all}, comparing the groundtruth trajectory with the trajectory estimate after aligning both trajectories regarding the position and yaw angle. 

\iftoggle{eccv}{

\begin{figure}
    \centering
    \iftoggle{cvpr}{
        \includegraphics[width=0.32\textwidth]{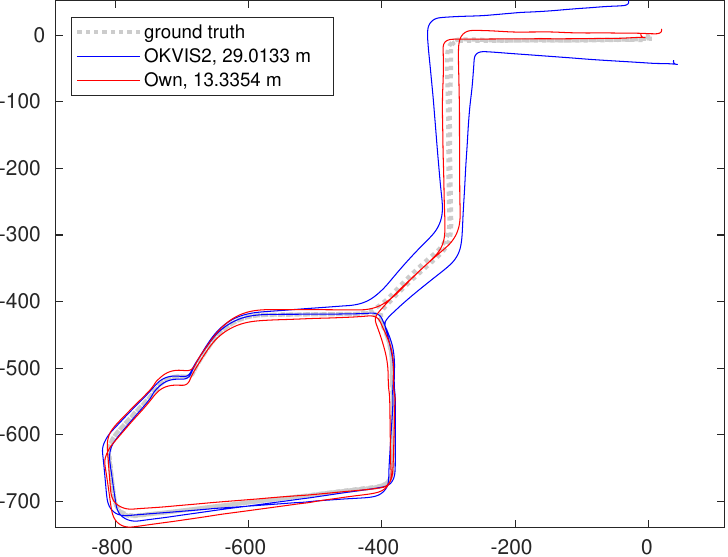}
        \includegraphics[width=0.32\textwidth]{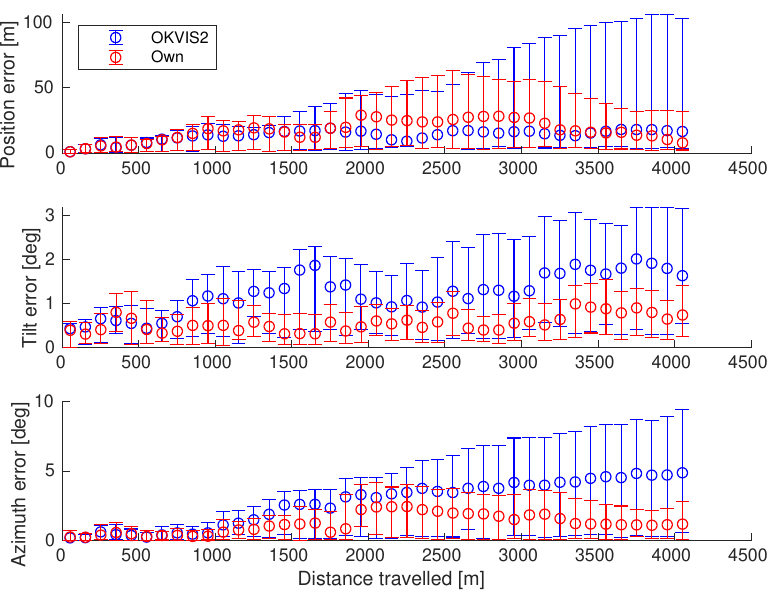}
        \caption{Trajectories and drift statistics of our method vs. the baseline OKVIS2~\cite{Leutenegger2022} on the test sequence  `2020-06-12\_10-10-57` of the TUM4Season dataset~\cite{wenzel2020fourseasons}. The relative motion errors for position and orientation are aggregated over different sub-trajectory lengths. Our method drifts significantly less than the baseline, reducing especially the azimuth error.}
    }{}
    \iftoggle{eccv}{
        \includegraphics[width=0.32\textwidth]{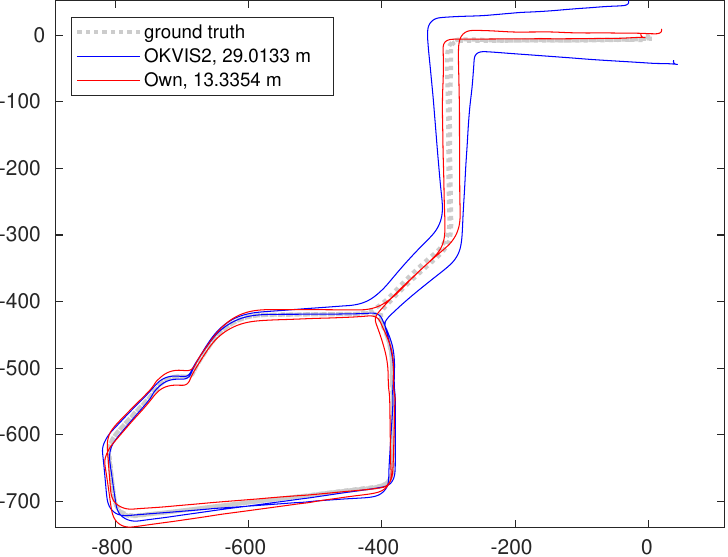}
        \includegraphics[width=0.32\textwidth]{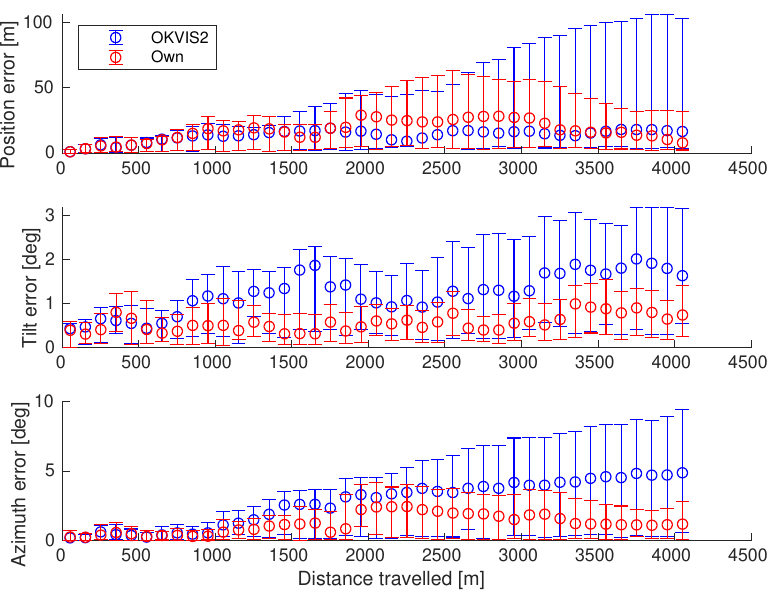}
        \includegraphics[width=0.32\textwidth]{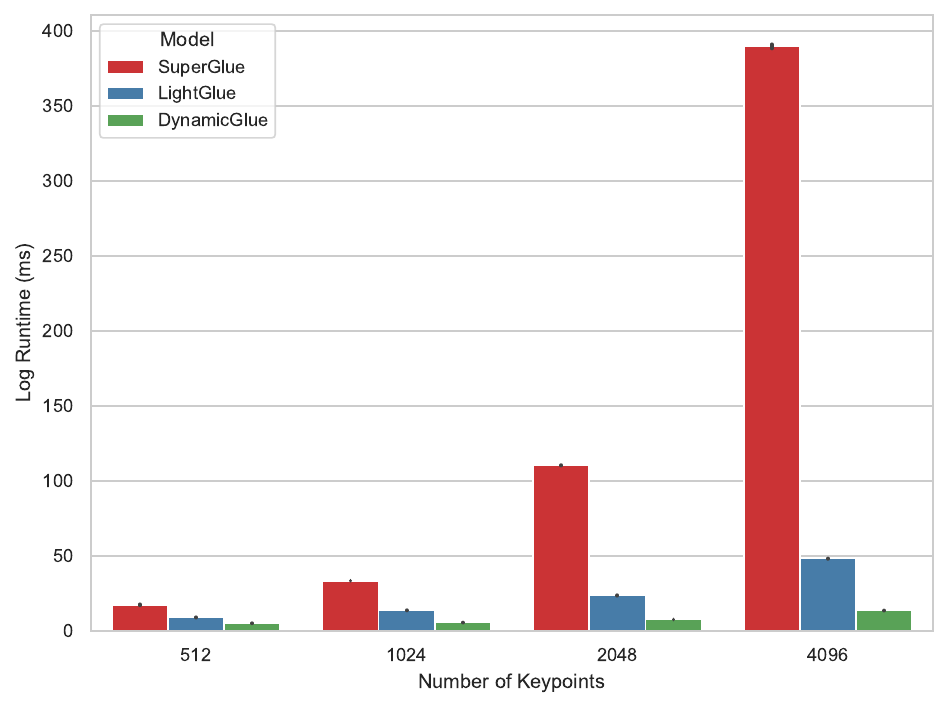}
        \caption{Left: Trajectories and drift statistics of our method vs. the baseline OKVIS2~\cite{Leutenegger2022} on the test sequence  `2020-06-12\_10-10-57` of the TUM4Season dataset~\cite{wenzel2020fourseasons}. The relative motion errors for position and orientation are aggregated over different sub-trajectory lengths. Our method drifts significantly less than the baseline, reducing especially the azimuth error. Right: Runtime comparison of our framework in comparison to Superglue~\cite{Sarlin2020} and LightGlue~\cite{Lindenberger2023}.}
    }{}
    \label{fig:okvis_trajectory}
    \vspace{-0.5cm}
\end{figure}
}{}
\iftoggle{corl}{

}{}

Especially in \ac{VIO} mode, our network enhances the system's overall performance quantified by the \ac{ATE} by 29\%. 
The most significant improvements are evident in the azimuth error in sessions with a medium or high number of moving objects, most likely due to the erroneous tracking of nearby moving objects in the original OKVIS2.
In \ac{SLAM} mode, OKVIS2 significantly improves its performance by leveraging loop closures such that both methods result in similar results. 
Therefore, our method is especially powerful in scenarios without revisited locations and long sequences where drift can accumulate. 
Remarkably, for high dynamic scenarios, employing our system in \ac{VIO} vastly outperforms these baselines, including the state-of-the-art matching framework LightGlue~\cite{Lindenberger2023}.
\section{Conclusion}
\label{sec:conclusion}
Most previous methods in keypoint matching focused on appearance changes, such as large viewpoint changes, which can lead to massive drift in downstream tasks, such as \ac{SLAM}.
In this work, we introduced \frameworkname, which is, to the best of our knowledge, the first to propose an additional awareness of the scene dynamics.
By augmenting a graph neural network with epipolar and temporal information, we achieve state-of-the-art matching performance and vastly outperform previous methods regarding dynamic scenes by $85\%$ while immensely reducing the required computational complexity by only including a small subset of neighboring keypoints into the graph.
Exemplary for downstream tasks, we demonstrate that our framework can lead to massive improvements in the accuracy of \ac{VIO} and \ac{SLAM} pipelines, leading to a $29\%$ decrease in \ac{ATE} over a state-of-the-art system in \ac{VIO} mode, and up to $43\%$ for medium to high dynamic scenes.
While we have shown good generalization capabilities of our model, with our proposed self-supervised training pipeline, the network can easily be fine-tuned to new environments without any human intervention.

\nottoggle{corl}{\begin{figure*}[t!]
    \centering
    \def\coloroverpic{lime}
    \def\sizeoverpic{\normalsize}
    \def\colorbbox{green}
    \def\thicknessbbox{0.5mm}

    \iftoggle{cvpr}{
    
    \begin{overpic}[width=0.48\textwidth,height=4.0cm]{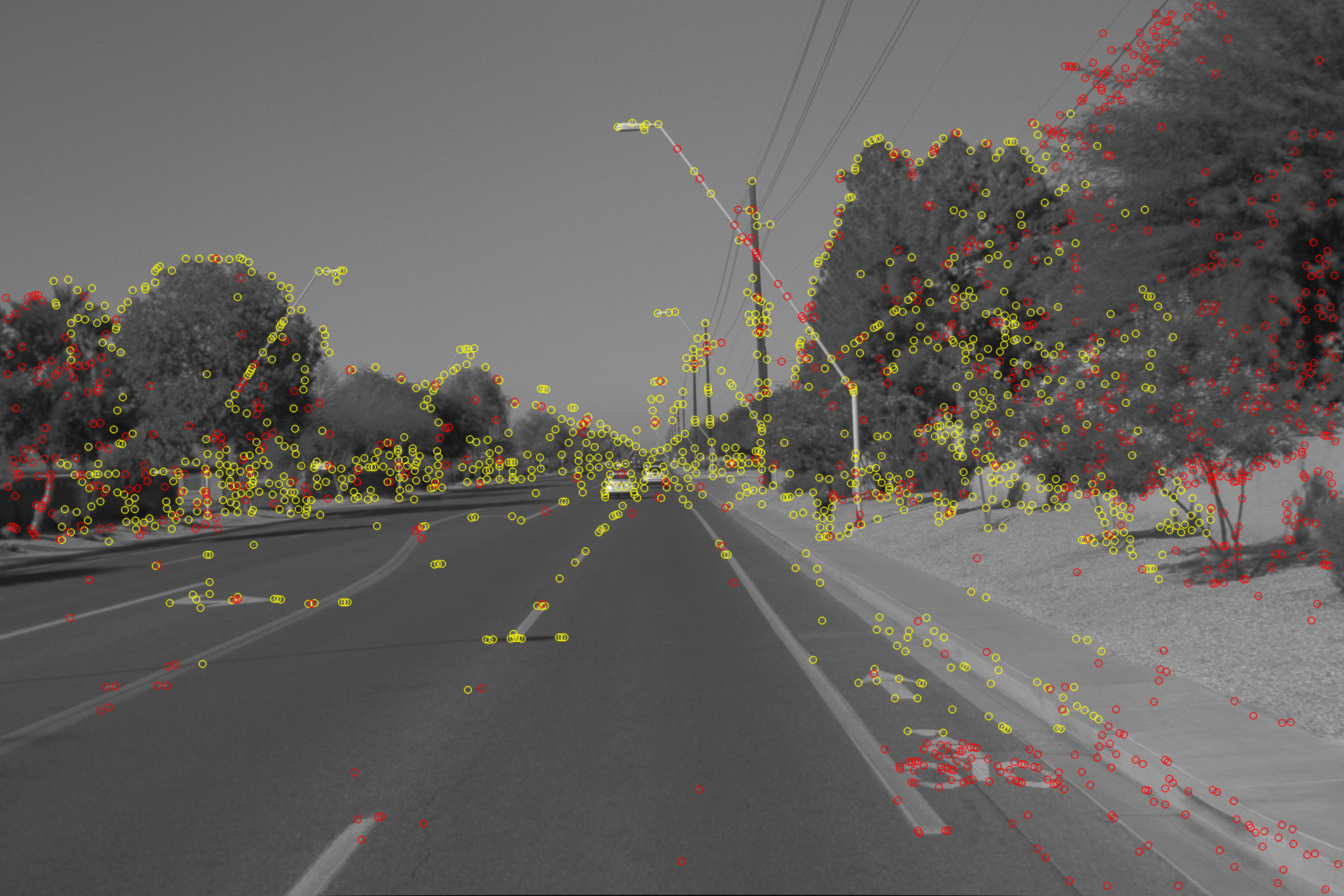}
        \put(3,40){\color{\coloroverpic}\sizeoverpic\textbf{LightGlue}}
        \put(0, 0){\color{\colorbbox}\linethickness{\thicknessbbox}\polygon(42, 18)(42, 27)(53, 27)(53, 18)}
    \end{overpic}
    \begin{overpic}[width=0.48\textwidth,height=4.0cm]{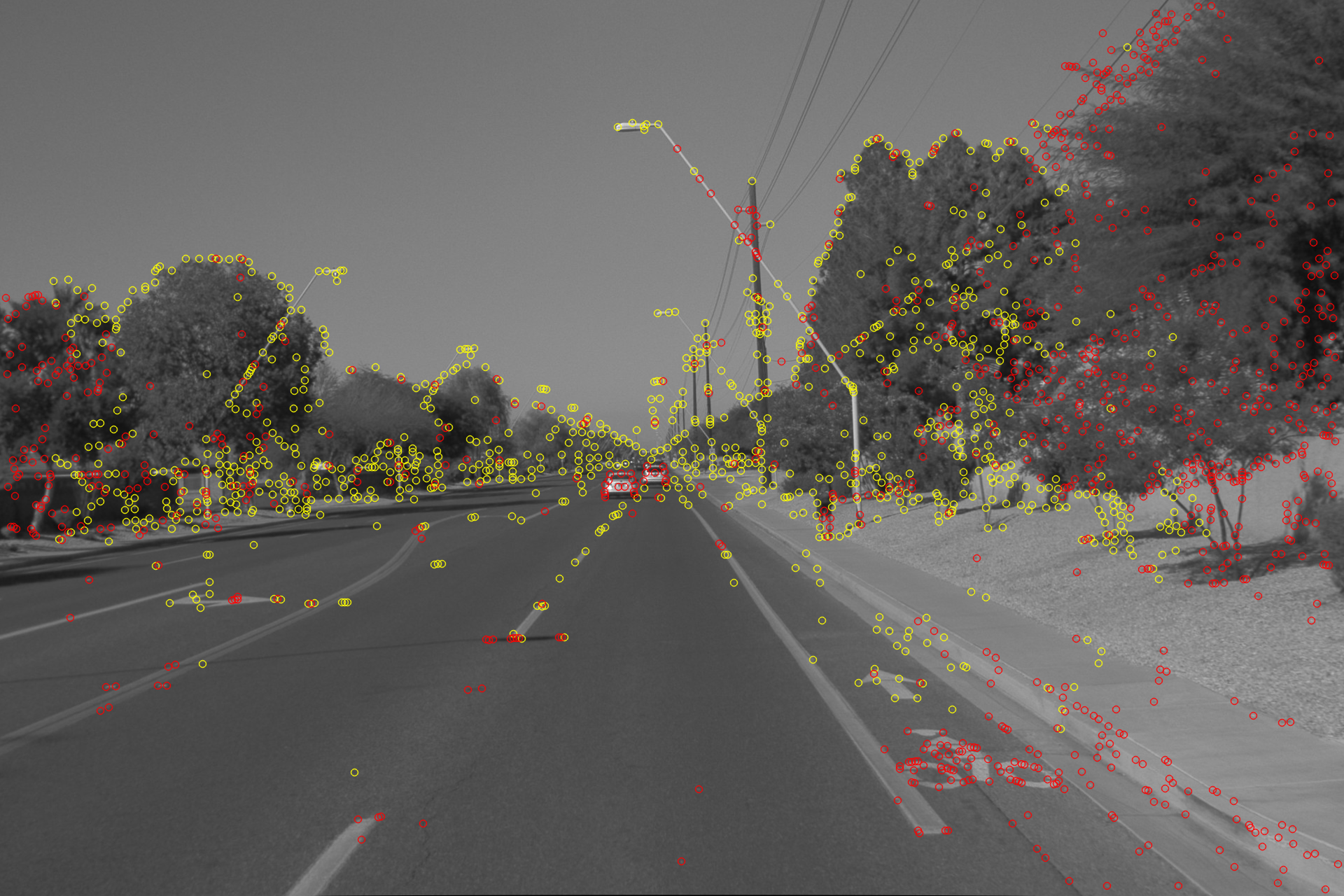}
        \put(3,40){\color{\coloroverpic}\sizeoverpic\textbf{\frameworkname}}
        \put(0, 0){\color{\colorbbox}\linethickness{\thicknessbbox}\polygon(42, 18)(42, 27)(53, 27)(53, 18)}
    \end{overpic}
    
    \begin{overpic}[width=0.48\textwidth,height=4.0cm]{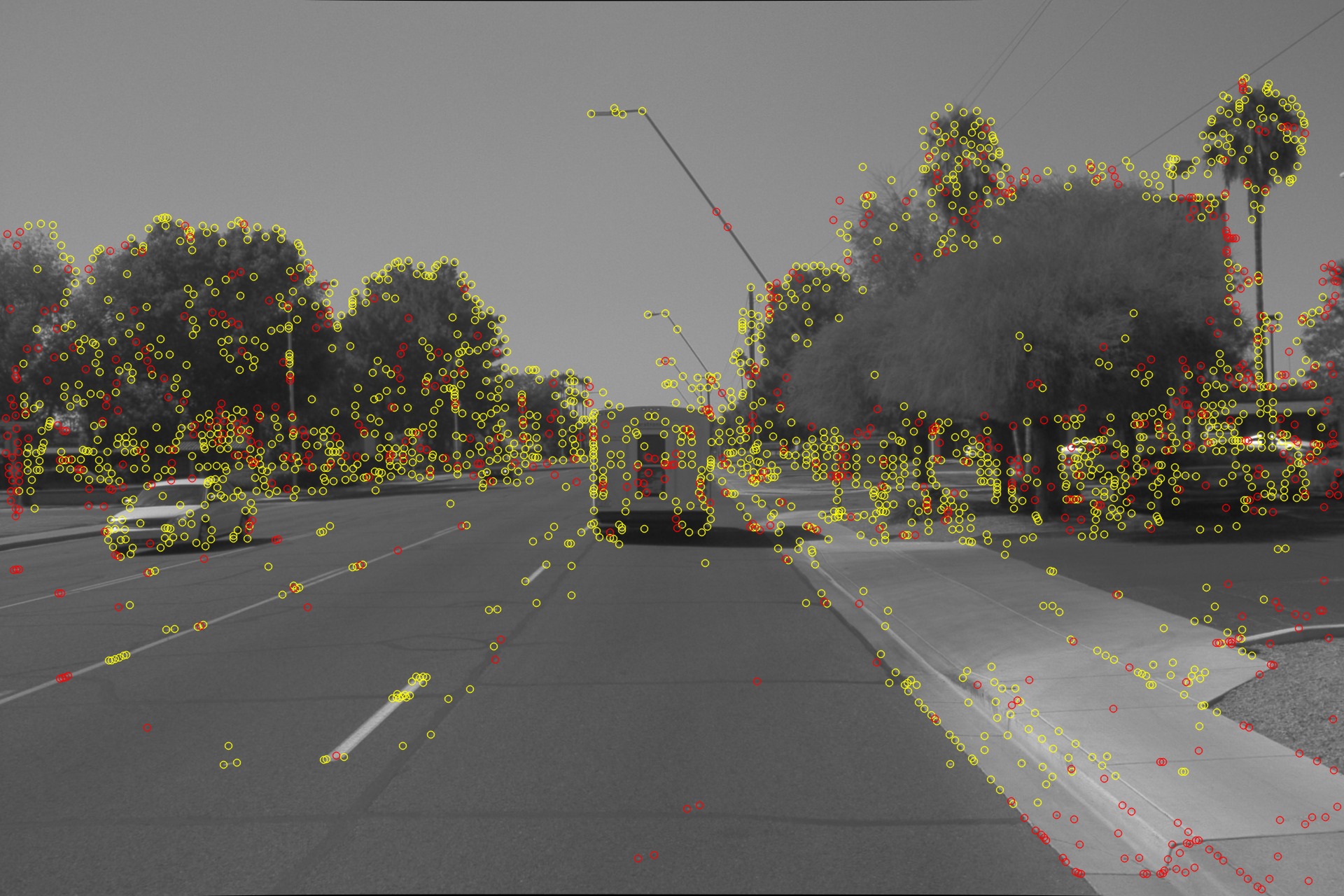}
        \put(3,40){\color{\coloroverpic}\sizeoverpic\textbf{LightGlue}}
        \put(0, 0){\color{\colorbbox}\linethickness{\thicknessbbox}\polygon(6, 16)(6, 25)(20, 25)(20, 16)}
    \end{overpic}
    \begin{overpic}[width=0.48\textwidth,height=4.0cm]{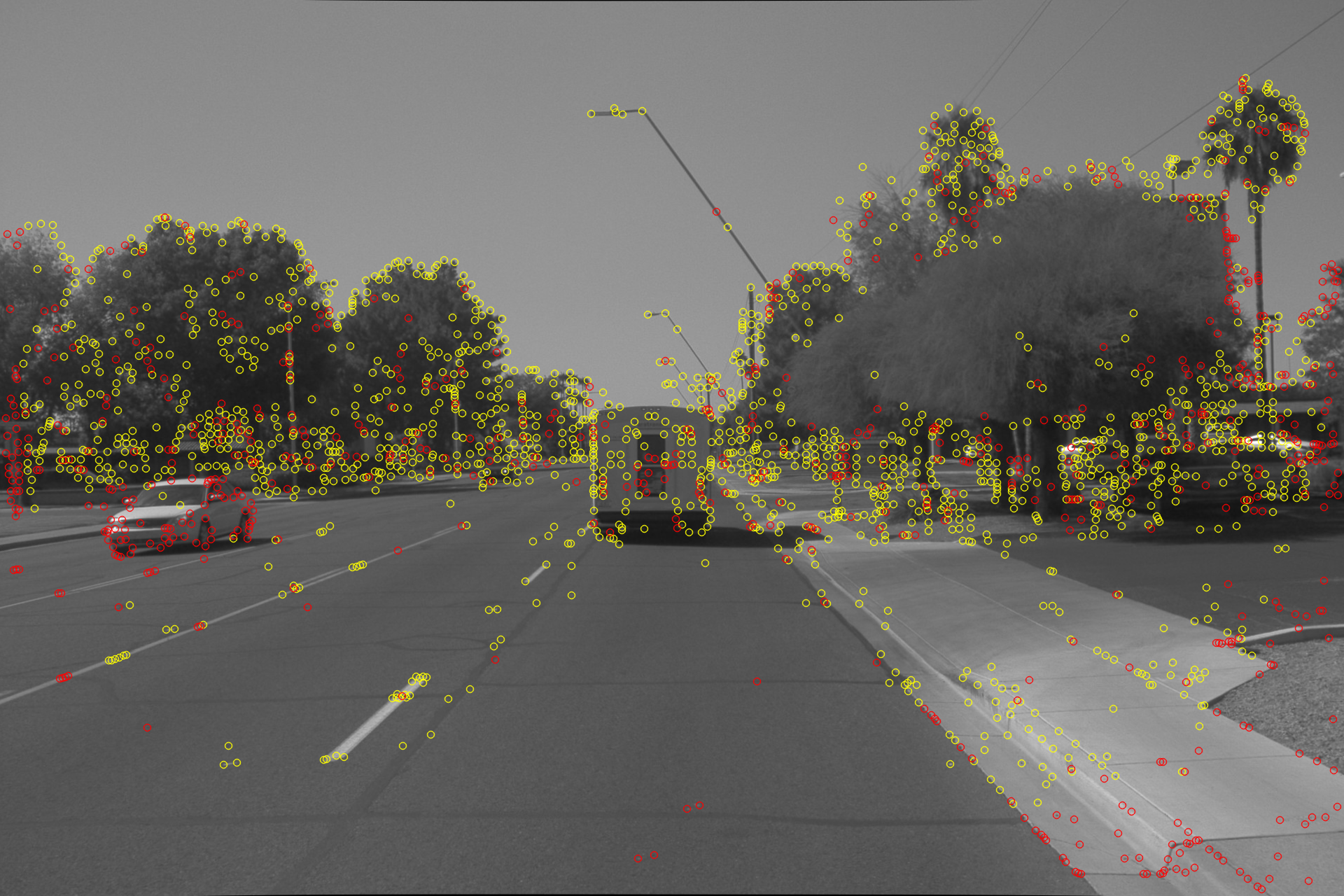}
        \put(3,40){\color{\coloroverpic}\sizeoverpic\textbf{\frameworkname}}
        \put(0, 0){\color{\colorbbox}\linethickness{\thicknessbbox}\polygon(6, 16)(6, 25)(20, 25)(20, 16)}
    \end{overpic}
    
    \begin{overpic}[width=0.48\textwidth,height=4.8cm]{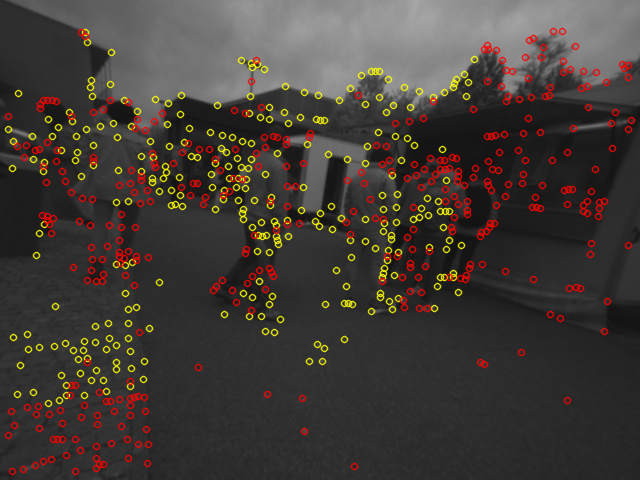}
        \put(3,50){\color{\coloroverpic}\sizeoverpic\textbf{LightGlue}}
        \put(0, 0){\color{\colorbbox}\linethickness{\thicknessbbox}\polygon(32, 20)(32, 43)(80, 43)(80, 20)}
    \end{overpic}
    \begin{overpic}[width=0.48\textwidth,height=4.8cm]{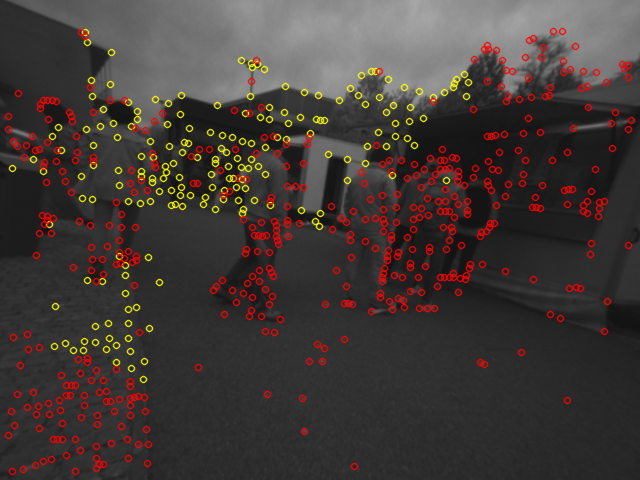}
        \put(3,50){\color{\coloroverpic}\sizeoverpic\textbf{\frameworkname}}
        \put(0, 0){\color{\colorbbox}\linethickness{\thicknessbbox}\polygon(32, 20)(32, 43)(80, 43)(80, 20)}
    \end{overpic}
    
    \begin{overpic}[width=0.48\textwidth,height=4.8cm]{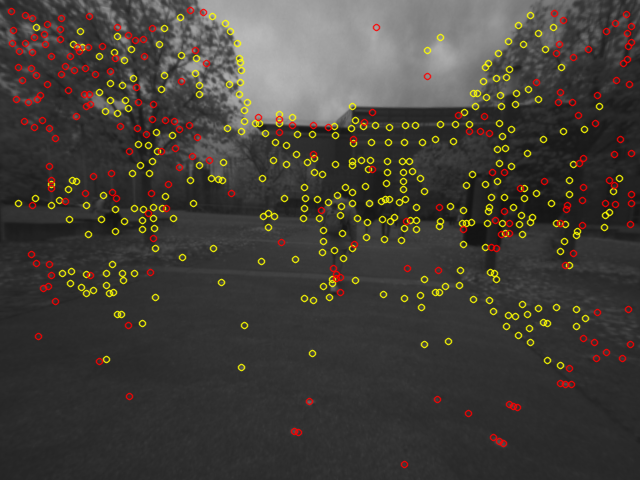}
        \put(3,50){\color{\coloroverpic}\sizeoverpic\textbf{LightGlue}}
        \put(0, 0){\color{\colorbbox}\linethickness{\thicknessbbox}\polygon(49, 20)(49, 37)(57, 37)(57, 20)}
    \end{overpic}
    \begin{overpic}[width=0.48\textwidth,height=4.8cm]{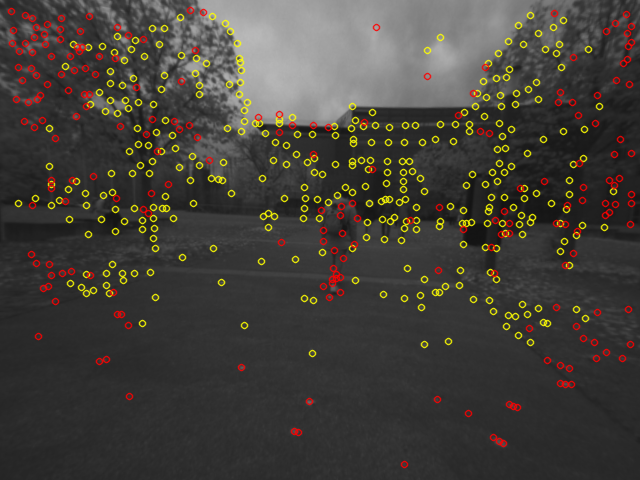}
        \put(3,50){\color{\coloroverpic}\sizeoverpic\textbf{\frameworkname}}
        \put(0, 0){\color{\colorbbox}\linethickness{\thicknessbbox}\polygon(49, 20)(49, 37)(57, 37)(57, 20)}
    \end{overpic}
    
    }{}
    \iftoggle{eccv}{

    \begin{overpic}[width=0.48\textwidth,height=4.0cm]{media/matching_qualitative/waymo/LightGlue/session_1482/375.png}
        \put(3,60){\color{\coloroverpic}\sizeoverpic\textbf{LightGlue}}
        \put(0, 0){\color{\colorbbox}\linethickness{\thicknessbbox}\polygon(42, 28)(42, 37)(53, 37)(53, 28)}
    \end{overpic}
    \begin{overpic}[width=0.48\textwidth,height=4.0cm]{media/matching_qualitative/waymo/ours/session_1482/375.png}
        \put(3,60){\color{\coloroverpic}\sizeoverpic\textbf{\frameworkname}}
        \put(0, 0){\color{\colorbbox}\linethickness{\thicknessbbox}\polygon(42, 28)(42, 37)(53, 37)(53, 28)}
    \end{overpic}
    
    \begin{overpic}[width=0.48\textwidth,height=4.0cm]{media/matching_qualitative/waymo/LightGlue/17160696560226550358_6229_820_6249_820/670.png}
        \put(3,60){\color{\coloroverpic}\sizeoverpic\textbf{LightGlue}}
        \put(0, 0){\color{\colorbbox}\linethickness{\thicknessbbox}\polygon(6, 23)(6, 35)(20, 35)(20, 23)}
    \end{overpic}
    \begin{overpic}[width=0.48\textwidth,height=4.0cm]{media/matching_qualitative/waymo/ours/17160696560226550358_6229_820_6249_820/670.png}
        \put(3,60){\color{\coloroverpic}\sizeoverpic\textbf{\frameworkname}}
        \put(0, 0){\color{\colorbbox}\linethickness{\thicknessbbox}\polygon(6, 23)(6, 35)(20, 35)(20, 23)}
    \end{overpic}
    
    \begin{overpic}[width=0.48\textwidth,height=4.8cm]{media/matching_qualitative/campus/LightGlue/02-11-2023-12-31-46/1385.png}
        \put(3,75){\color{\coloroverpic}\sizeoverpic\textbf{LightGlue}}
        \put(0, 0){\color{\colorbbox}\linethickness{\thicknessbbox}\polygon(32, 25)(32, 60)(80, 60)(80, 25)}
    \end{overpic}
    \begin{overpic}[width=0.48\textwidth,height=4.8cm]{media/matching_qualitative/campus/ours/02-11-2023-12-31-46/1385.png}
        \put(3,75){\color{\coloroverpic}\sizeoverpic\textbf{\frameworkname}}
        \put(0, 0){\color{\colorbbox}\linethickness{\thicknessbbox}\polygon(32, 25)(32, 60)(80, 60)(80, 25)}
    \end{overpic}
    
    \begin{overpic}[width=0.48\textwidth,height=4.8cm]{media/matching_qualitative/campus/LightGlue/02-11-2023-12-23-55/1175.png}
        \put(3,75){\color{\coloroverpic}\sizeoverpic\textbf{LightGlue}}
        \put(0, 0){\color{\colorbbox}\linethickness{\thicknessbbox}\polygon(45, 27)(45, 53)(60, 53)(60, 27)}
    \end{overpic}
    \begin{overpic}[width=0.48\textwidth,height=4.8cm]{media/matching_qualitative/campus/ours/02-11-2023-12-23-55/1175.png}
        \put(3,75){\color{\coloroverpic}\sizeoverpic\textbf{\frameworkname}}
        \put(0, 0){\color{\colorbbox}\linethickness{\thicknessbbox}\polygon(45, 27)(45, 53)(60, 53)(60, 27)}
    \end{overpic}
    
    }{}

    \iftoggle{corl}{

    \begin{overpic}[width=0.48\textwidth,height=4.0cm]{media/matching_qualitative/waymo/LightGlue/session_1482/375.png}
        \put(3,53){\color{\coloroverpic}\sizeoverpic\textbf{LightGlue}}
        \put(0, 0){\color{\colorbbox}\linethickness{\thicknessbbox}\polygon(42, 28)(42, 37)(53, 37)(53, 28)}
    \end{overpic}
    \begin{overpic}[width=0.48\textwidth,height=4.0cm]{media/matching_qualitative/waymo/ours/session_1482/375.png}
        \put(3,53){\color{\coloroverpic}\sizeoverpic\textbf{\frameworkname}}
        \put(0, 0){\color{\colorbbox}\linethickness{\thicknessbbox}\polygon(42, 28)(42, 37)(53, 37)(53, 28)}
    \end{overpic}
    
    \begin{overpic}[width=0.48\textwidth,height=4.0cm]{media/matching_qualitative/waymo/LightGlue/17160696560226550358_6229_820_6249_820/670.png}
        \put(3,53){\color{\coloroverpic}\sizeoverpic\textbf{LightGlue}}
        \put(0, 0){\color{\colorbbox}\linethickness{\thicknessbbox}\polygon(6, 23)(6, 35)(20, 35)(20, 23)}
    \end{overpic}
    \begin{overpic}[width=0.48\textwidth,height=4.0cm]{media/matching_qualitative/waymo/ours/17160696560226550358_6229_820_6249_820/670.png}
        \put(3,53){\color{\coloroverpic}\sizeoverpic\textbf{\frameworkname}}
        \put(0, 0){\color{\colorbbox}\linethickness{\thicknessbbox}\polygon(6, 23)(6, 35)(20, 35)(20, 23)}
    \end{overpic}
    
    \begin{overpic}[width=0.48\textwidth,height=4.8cm]{media/matching_qualitative/campus/LightGlue/02-11-2023-12-31-46/1385.png}
        \put(3,65){\color{\coloroverpic}\sizeoverpic\textbf{LightGlue}}
        \put(0, 0){\color{\colorbbox}\linethickness{\thicknessbbox}\polygon(32, 25)(32, 60)(80, 60)(80, 25)}
    \end{overpic}
    \begin{overpic}[width=0.48\textwidth,height=4.8cm]{media/matching_qualitative/campus/ours/02-11-2023-12-31-46/1385.png}
        \put(3,65){\color{\coloroverpic}\sizeoverpic\textbf{\frameworkname}}
        \put(0, 0){\color{\colorbbox}\linethickness{\thicknessbbox}\polygon(32, 25)(32, 60)(80, 60)(80, 25)}
    \end{overpic}
    
    \begin{overpic}[width=0.48\textwidth,height=4.8cm]{media/matching_qualitative/campus/LightGlue/02-11-2023-12-23-55/1175.png}
        \put(3,65){\color{\coloroverpic}\sizeoverpic\textbf{LightGlue}}
        \put(0, 0){\color{\colorbbox}\linethickness{\thicknessbbox}\polygon(45, 27)(45, 53)(60, 53)(60, 27)}
    \end{overpic}
    \begin{overpic}[width=0.48\textwidth,height=4.8cm]{media/matching_qualitative/campus/ours/02-11-2023-12-23-55/1175.png}
        \put(3,65){\color{\coloroverpic}\sizeoverpic\textbf{\frameworkname}}
        \put(0, 0){\color{\colorbbox}\linethickness{\thicknessbbox}\polygon(45, 27)(45, 53)(60, 53)(60, 27)}
    \end{overpic}
    
    }{}
    
    \caption{Qualitative results of our framework (right) in various scenarios compared to LightGlue~\cite{Lindenberger2023} (left). Matched keypoints are shown in yellow, unmatched in red. Our method can distinguish dynamic from static objects in different environments and diverse object types.}
    \label{fig:qualitative_examples}
    \vspace{-0.4cm}
\end{figure*}
\clearpage}{}

\clearpage
{
    \iftoggle{cvpr}{
        \small
        \bibliographystyle{ieeenat_fullname}    
    }{}
    \iftoggle{eccv}{
        \bibliographystyle{splncs04}       
    }{}
    \bibliography{main}
}


\end{document}